%% file: main.tex
\PassOptionsToPackage{pdftex, xetex}{graphicx}
\PassOptionsToPackage{usenames,dvipsnames,svgnames,table}{xcolor}
\documentclass[letterpaper, 10 pt, conference]{ieeeconf}
\IEEEoverridecommandlockouts
\makeatletter
\let\NAT@parse\undefined
\makeatother
\usepackage[sort,numbers,compress]{natbib}

\usepackage{pratik}
\input{notation}
\graphicspath{{./figures42/}}

\usepackage{algpseudocode}
\usepackage{listings}

\makeatletter
\DeclareMathSizes{\@xpt}{9}{7}{5}
\makeatother

\title{Active Perception using Neural Radiance Fields}

\author{Siming He, Christopher D. Hsu$^{*}$, Dexter Ong$^{*}$, Yifei Simon Shao, Pratik Chaudhari
\thanks{$^*$Equal Contribution}
\thanks{
General Robotics, Automation, Sensing and Perception (GRASP) Laboratory at the University of Pennsylvania.
\href{mailto:siminghe@seas.upenn.edu}{siminghe@seas.upenn.edu}
\href{mailto:chsu8@seas.upenn.edu}{chsu8@seas.upenn.edu},
\href{mailto:odexter@seas.upenn.edu}{odexter@seas.upenn.edu},
\href{mailto:yishao@seas.upenn.edu}{yishao@seas.upenn.edu},
\href{mailto:pratikac@seas.upenn.edu}{pratikac@seas.upenn.edu}}
}
\begin{document}


\maketitle
\thispagestyle{empty}

\begin{abstract}
We study active perception from first principles to argue that an autonomous agent performing active perception should maximize the mutual information that past observations posses about future ones. Doing so requires (a) a representation of the scene that summarizes past observations and the ability to update this representation to incorporate new observations (state estimation and mapping), (b) the ability to synthesize new observations of the scene (a generative model), and (c) the ability to select control trajectories that maximize predictive information (planning). This motivates a neural radiance field (NeRF)-like representation which captures photometric, geometric and semantic properties of the scene grounded. This representation is well-suited to synthesizing new observations from different viewpoints. And thereby, a sampling-based planner can be used to calculate the predictive information from synthetic observations along dynamically-feasible trajectories. We use active perception for exploring cluttered indoor environments and employ a notion of semantic uncertainty to check for the successful completion of an exploration task. We demonstrate these ideas via simulation in realistic 3D indoor environments\footnote{Code is available at \href{https://github.com/grasp-lyrl/Active-Perception-using-Neural-Radiance-Fields}{https://github.com/grasp-lyrl/Active-Perception-using-Neural-Radiance-Fields}. Full manuscript is available at \href{https://arxiv.org/abs/2310.09892}{https://arxiv.org/abs/2310.09892}.}.
\end{abstract}

\input{intro}
\input{problem}
\input{method}
\input{experiments}
\input{discussion}
\input{acknowledgments}

\begin{footnotesize}
\bibliographystyle{ieeetr}
\bibliography{references,pratik}
\end{footnotesize}

\end{document}

%% file: notation.tex
\def \KL {\text{KL}}
\def \I {\text{I}}
\def \Ipred {\text{I}_{\text{pred}}}
\def \SE {\text{SE}}

\def \S {\text{S}}

\def \ytp {y_{\text{all}}}
\def \yt {y_{\text{past}}}
\def \yttp {y_{\text{future}}}

\def \uttp {u_{\text{future}}}

%% file: intro.tex

\section{Introduction}
\label{s:intro}

The goal of a firefighter who is exploring a house after an earthquake is to locate all objects of interest, e.g., a person who might be in distress, the kitchen may be on fire etc. Such a search must be done quickly, in a new environment often in the absence of a map (if a part of the house has collapsed, the old map is not very useful\dots). If there is no situation that requires immediate attention then everything is fine in this house, the firefighter should move on to the next one. Concluding that ``everything is fine'' is not very easy, a child might be trapped under the bed---one needs to search the entire house before declaring that everything is fine. Often, the firefighter might coarsely scan unseen areas and re-visit them later to obtain more details. This is a canonical search and rescue problem and effective solutions to this problem hold nuggets of insight into many other problems where an agent (human or robotic) explores an environment to discover information specific to a task. Our goal in this paper will be to formalize such problems using ideas from active perception.

\begin{figure}
    \centering
    \includegraphics[width=0.75\linewidth]{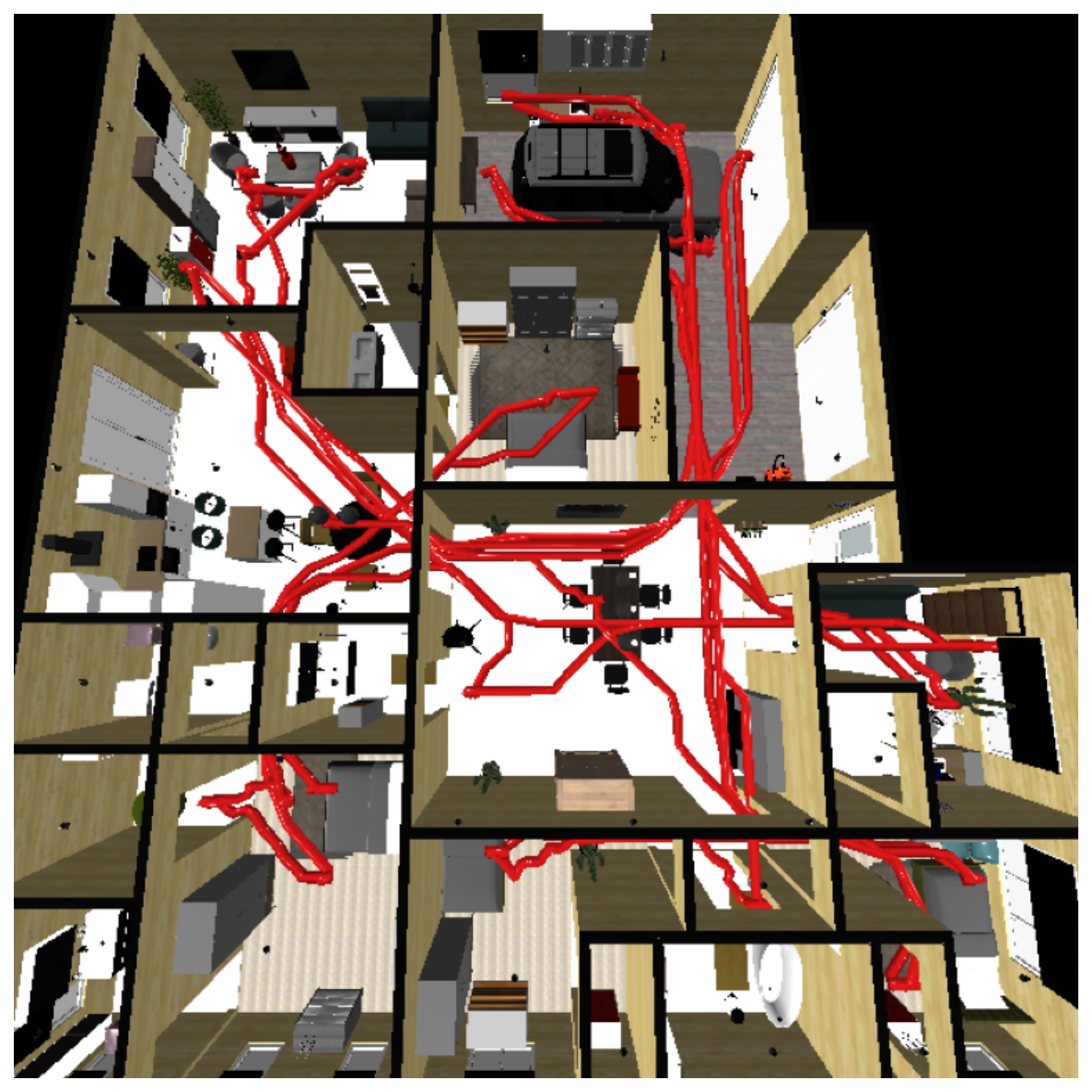}
    \includegraphics[width=\linewidth]{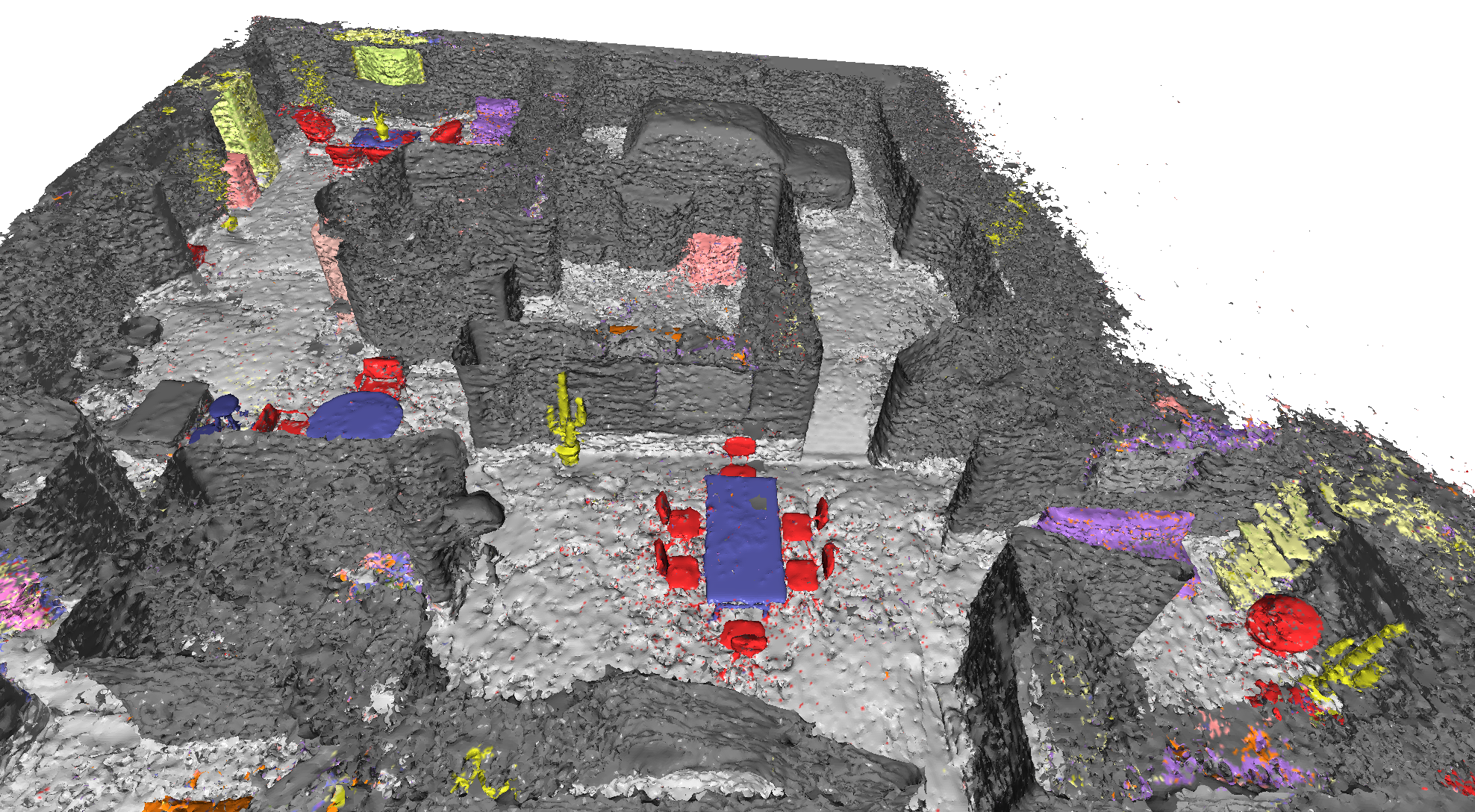}
    \caption{\textbf{Top:} Trajectories of a quadrotor that actively explores a complex and cluttered indoor environment to localize all the different kinds of objects. Our approach to active perception maximizes the mutual information of the past observations (RGBD images and semantic segmentation masks) with respect to future observations using a generative model to select highly informative trajectories that explore large parts of the scene quickly.
    \textbf{Bottom:} We build a neural-radiance field (NeRF) representation of the scene to calculate this mutual information. This provides us with an accurate representation of the free space within which we can sample dynamically-feasible trajectories for a differentially-flat model of a quadrotor. This picture shows a mesh constructed from the voxel-grid representation implicit inside the NeRF after active exploration; color denotes objects of different categories predicted by our semantic NeRF.
    }
    \label{fig:Overview}
\end{figure}

In a paper titled “Revisiting active perception”~\cite{bajcsy2016revisiting}, Bajcsy, Aloimonos, and Tsotsos defined active perception as \emph{``An agent is an active perceiver if it knows why it wishes to sense, and then chooses what to perceive, and determines how, when and where to achieve that perception.''}
Different fields have instantiated these ideas in different ways, e.g., to recover geometric properties of the scene in computer vision~\cite{Aloimonos2004ActiveV}, identifying these principles in biological perception systems~\cite{Tsotsos2007AttentionAV}, anticipating the environment to perform robotics tasks with low latency~\cite{scaramuzza2017quadactivevision}, searching for objects actively~\cite{novkovic2020object} etc. Fields outside robotics such as active learning and reinforcement learning~\cite{dosovitskiy2016learning} have formulated active perception as learning a statistic of past data that can be used to take actions at test time. These fields implement active perception by selecting primitives for (why, what, how, when and where) in a problem-dependent ad hoc fashion. This paper also focuses on one specific problem: exploration tasks where a robot seeks to find all objects in the scene. \textbf{We formalize active perception from first principles and distill this formulation down into the \emph{necessary} ingredients that an active perceiver should have.}

We argue that a \textbf{neural radiance field (NeRF)}~\cite{mildenhall2020nerf} representation is particularly well-suited to active perception tasks. It summarizes multi-modal information in a consistent fashion, e.g., the rendering equation ties together \textbf{geometry and photometry}, object segmentation and boundaries can be used to understand \textbf{semantics of the scene}~\cite{inplaceICCV2021}. A NeRF can also \textbf{synthesize new observations from the scene} across all these three modalities---and we will argue that this is a \textbf{necessary ingredient of active perception}. We exploit recent techniques that have made NeRFs computationally tractable~\cite{mueller2022instant,li2023nerfacc}. This allows us to \textbf{implement the entire formulation} of this paper, from processing input images, to selecting dynamically feasible trajectories in free space that maximize predictive information, and computing dense scene segmentation \textbf{only about 10$\times$ slower than real-time, even if we are using Python}, as shown in \cref{fig:Overview}.

We use ideas from information theory to \textbf{formulate active perception as maximizing the predictive information that past observations contain about future ones}~\cite{bialek2001predictability}. When coupled with a rich representation like a NeRF and a planning framework that can quickly plan collision-free dynamically-feasible trajectories in free space (also obtained from the same NeRF), we can achieve behaviors that are more sophisticated than other instantiations of active perception such as selecting the next best view~\cite{pan2022activenerf}, searching for single objects~\cite{smith2022uncertaintydriven}, maximizing/reducing the uncertainty of unknown parts of the scene or frontiers~\cite{yamauchi1997frontier, chaplot2020learning}, using heuristic based on information gain for navigation~\cite{kopanas2023improving}, using open-loop but learned exploration policies~\cite{marza2023autonerf}, computing the scene representation and the controller independently~\cite{adamkiewicz2022visiononly}, and optimizing trajectories between two fixed viewpoints~\cite{zhan2022activermap,Ran_2023}. The \textbf{central tenet of active perception} is to take control that leads to informative observations, and use these observations to calculate the next set of controls. This requires a \textbf{tightly integrated feedback loop between perception and control}. And although one often sees procedures that break this feedback for the sake of computational tractability, approaches like ours that \textbf{adhere faithfully to the active perception paradigm} should be more effective.

This formulation also suggests a natural mechanism to compute mutual information/uncertainty, namely using bootstrapped ensembles. Our approach can utilize other techniques to compute uncertainty, e.g.,~\cite{shen2021stochastic} but these methods require major modifications to the vanilla NeRF architecture and cannot exploit efficient implementations such as multi-resolution hash encoding. \cite{jiang2023fisherrf} uses Fisher information as a measure of information gain. \cite{sunderhauf2022densityaware} uses variance of RGB images from an ensemble of models and the density of rays to quantify uncertainty. We also \textbf{use ensembles to calculate a probability distribution over color, pixel depth, voxel occupancy and labels of objects}. But our mathematical formulation suggests that an active perception agent should maximize predictive information instead of uncertainty.

%% file: problem.tex

\section{Problem Formulation}
\label{s:problem}

Let $\xi$ represents a fixed scene from which we obtain a sequence of observations $y_1^t = (y_1, \ldots, y_{t-1})$ from viewpoints $x_1^t = (x_1,\ldots, x_{t-1})$ respectively where each $x_k \equiv (R_k, t_k) \in \SE(3)$. Let $y_t^{t+\Delta t}$ be the sequence of future observations for some $\Delta t \in \naturals_+$. We will use the shorthand
\[
    \aed{
        \yt \equiv y_1^t, \quad
        \yttp \equiv y_t^{t + \Delta t}, \quad
        \ytp \equiv y_1^{t + \Delta t}.
    }
\]
At each time step, we may write $p(y_t \mid \xi, x_t)$ as the probability of obtaining an observation $y_t$ from a viewpoint $x_t$. Assume that the quadrotor takes controls $u_t$ to go from $x_t$ to another $x_{t+1} = f(x_t, u_t)$. If we did not observe the past, then the future is drawn from a probability distribution $p(\yttp)$; past observations tell us that the future will be drawn from $p(\yttp \mid \yt)$. The gap between these two distributions is what one might call ``predictive information'', i.e., the amount of the information past observations contain about the future~\cite{bialek2001predictability}. We can write it as the mutual information:
\beq{
    \aed{
        \I(\yttp, \yt)
        &=\int \dd{p(\yt)} \KL \rbr{\yttp \mid \yt,\ \yttp},\\
        &=\int \dd{p(\yt)} \rbr{\S(\yttp) - \S(\yttp \mid \yt)},\\
        &= \S(\yt) + \S(\yttp) - \S(\ytp),
    }
    \label{eq:mi}
}
where $\S(\yt) = -\int \dd{p(\yt)} \log p(\yt)$ is the Shannon entropy of $\yt$ and the Kullback-Leibler (KL) divergence is $\KL(\yt, \yt') = \S(\yt, \yt') - \S(\yt)$.

Active exploration seeks to obtain information about the scene using the control authority of the agent. We can formalize active perception as selecting control sequences $\uttp$ to maximize the predictive information over some future time horizon of length $\Delta t$
\beq{
    \hat u_{\text{future}} \in \argmax_{\uttp} \I(\yttp, \yt).
    \label{eq:problem}
}
Future observations $\yttp$ depend upon the future control actions $\uttp$ and so does the mutual information objective.

\subsection{Predictive information characterizes the complexity of the scene in active perception}
\label{predinfo}

Predictive information can characterize the complexity of the scene and that is why maximizing it could enable active perception. We explain this using a few examples where $\Delta t \to \infty$. For ergodic observations $y_1^t$, the asymptotic equipartition theorem~\cite[Theorem 3.1.1]{cover1999elements} states that $\lim_{t \to \infty} \S(\yt)/t = \S(y \mid \xi)$. In general, entropy is an extensive quantity, i.e., it scales with the length of the sequence, so let us set $\S(\yt) \equiv t \S(y \mid \xi) + \S_1(t)$ with $\S_1(t)$ being the non-extensive part. Using~\cref{eq:mi}, we can now define
\[
    \Ipred(t) \doteq \lim_{\Delta t \to \infty} \I(\yttp, \yt) = \S_1(\yt),
\]
as the predictive information of an infinite future sequence contained in the past. To get the above expression, we have used the fact that $\lim_{\Delta t \to \infty} \S_1(\yttp) = \S_1(\ytp)$. If we know the scene and the sequence $\yt$ is ergodic, then $\Ipred(t)$ is a constant. This is because under these conditions, for any $\e > 0$, we can find a large enough $t$ such that the set of typical sequences $\yt$ has probability greater than $1 -\e$. No matter how long we observe, we only obtain a finite amount of information about the future, e.g., a sequence of states from a Markov chain with known transition matrix, or noisy observations from a perfectly known scene $\xi$ due to stochasticity in the rendering process. Such situations are not interesting for active perception.

Active perception is non-trivial when the predictive information diverges $\lim_{t \to \infty} \Ipred(t) = \infty$. The authors in~\cite{bialek2001predictability} study two such regimes. When the model (scene $\xi$ for us) is finite-dimensional with $\l$ parameters, they show that $\S_1(t) \simeq \f{\l}{2} \log t +\OO(1)$ as $t \to \infty$. In this case, each successive observation gives us additional information and predictive information is close to the Bayesian Information Criterion (BIC). If the model has an infinite number of parameters, then $\Ipred$ can grow more quickly, like a power law $\S_1(t) \simeq t^\nu$ for some constant $\nu < 1$ (since $\S_1$ has to be non-extensive). In this case, we discover increasingly finer details of the scene with more and more observations.

Observations of trivial scenes (e.g., empty room) have constant $\Ipred$; observations of finite-dimensional scenes (e.g., non-textured surfaces, finite number of objects, etc.) have $\Ipred$ that grows logarithmically with $t$; and observations of complex scenes (e.g., textured surfaces, unbounded scenes) give us $\Ipred$ that grows more quickly, say a power law. The second setting is relevant for searching a room to find an object; the third is akin to the ``spot the difference'' game where one searches for differences in two pictures.

\subsection{Instantiating the predictive information objective}

We can rewrite~\cref{eq:mi} as \beq{
     \I(\yttp, \yt) = \S(\yttp) - \int \dd{p(\yt)} \S(\yttp \mid \yt).
     \label{eq:pred_i}
}

\textbf{Postulate 1:} The true scene $\xi$ is a sufficient statistic of the past observations $\yt$ for predicting the future ones $\yttp$. We will assume that the viewpoint $x_t$ is known at all times $t$ and thereby suppress $p(y \mid \xi, g)$ to simply $p(y \mid \xi)$. Then
\beqs{
    \int \dd{p(\yt)} \S(\yttp \mid \yt) = \int \dd{p(\xi)} \S(\yttp \mid \xi).
    \label{eq:S_yttp_yt}
}
We can also write
\beqs{
    \aed{
    p(\yttp) = \int \dd{p(\yt)} p(\yttp \mid \yt)
    =\int \dd{p(\xi)} p(\yttp \mid \xi).
    }
    \label{eq:S_yttp}
}
The first term $\S(\yttp)$ is the entropy of the marginalized future over possible scenes $\xi$, while the second term $\int \dd{p(\xi)} \S(\yttp \mid \xi)$ is the entropy of the future averaged over these scenes. Suppose we build
\beq{
    p(\xi) = \f{1}{m} \sum_{k=1}^m \delta_{\xi_k}(\xi)
    \label{eq:p_xi}
}
using an ensemble of $m$ scenes $\{\xi_k\}_{k=1}^m$, here $\delta$ is the Dirac delta distribution. Then we can instantiate the predictive information objective by noting that the second term is the average uncertainty over the future predicted by each of these scenes and the first term is the uncertainty over the future of the joint distribution of their predictions.

For $\xi$ to be a sufficient statistic of the past observations for predicting future ones, it should correspond to some kind of a ``map''. For a computationally efficient implementation, we \emph{should} update this map/statistic incrementally instead of estimating it from $\yt$ every time. We have therefore asserted the need of mapping and state estimation respectively.

\textbf{Postulate 2:} We need to learn a generative model $p(\yttp \mid \xi)$ to calculate terms like $\S(\yttp \mid \xi)$. There are many ways to build such a model. For example, if we can calculate $p(\yttp \mid \xi)$ and sample $N$ sequences $\{y^{(k)}_{\text{future}}\}_{k=1}^N$, then
\beq{
    \S(\yttp \mid \xi) \approx -\f{1}{N} \sum_{k=1}^N p(y^{(k)}_{\text{future}} \mid \xi)\ \log p(y^{(k)}_{\text{future}} \mid \xi).
    \label{eq:S_monte_carlo}
}
The implementation of the generative model depends upon the problem. We are interested in tasks such as localizing different objects in the scene which require future sequences to be scored correctly in terms of their photometric, geometric and semantic consistency. There are many recent techniques that satisfy these desiderata piecemeal, e.g., a variational auto-encoder (VAE) or a diffusion model can generate photometrically accurate images but they do not model geometry; a GAN does not give access to a likelihood; a segmented point-cloud does not represent photometric properties of objects. NeRFs represent photometric and geometric properties well but they do not represent the semantics. In~\cref{s:method} we will use a ``semantic variant'' of NeRF as our generative model for predictive information.

The need for a generative model is not an ``assumption'', it is a direct outcome of our definition of active perception. Without the ability to synthesize or hallucinate new observations from the scene, we cannot perform active perception.

\textbf{Postulate 3:} We need a mechanism to select controls that maximize the predictive information. Future observations $\yttp$ are a function of controls $\uttp$. The active perception objective can be written to include the viewpoint as
\[
    \aed{
        \argmax\ \S(y_t^{t +\D t} \mid x_t^{t + \D t}) - \int \dd{p(\xi)} \S(y_t^{t +\D t} \mid \xi, x_t^{t + \D t})
    }
\]
with $x_{t+1} = f(x_t, u_t)$ for all $t$. 

\begin{example}[Linear system]
Suppose $x_t \in \reals^d$, and $f$ is an affine function, i.e., $x_{t+1} = A x_t + B u_t$ is the dynamics with $A \in \reals^{d \times d}$, $B \in \reals^d$, $u_t \in \reals$, and $\reals \ni y_t = \xi^\top x_t + \s_\w \w_t$ for zero-mean unit variance Gaussian white noise $\w \in \reals$ are the observations from each viewpoint. The ``scene'' $\xi \in \reals^d$ is the only unknown parameter. We could estimate $p(\xi \mid y_1^t) = \text{N}(\mu_\xi, \Sigma_\xi)$ using a Kalman filter. The second term $\S(y_t^{t +\D t} \mid \xi, x_t^{t + \D t})$ does not depend on $u_t^{t+\D t}$. Observe that
\[
    \S(y_t^{t+\D t} \mid x_t^{t+ \D t}) = \sum_{s=t}^{t+\D t} \S(y_s \mid x_s),
\]
where $p(y_s \mid x_s) = \text{N}(\mu_\xi^\top x_s, x_s^\top \Sigma_\xi x_s)$. Effectively, the predictive information objective is
\beq{
    \textstyle \argmax_{u_t^{t+\D t}}\ \sum_{s=t}^{t+\D t} \log \det \rbr{x_s^\top \Sigma_\xi x_s}.
    \label{eq:linear_setting}
}
It is maximized by $x_s$ along the eigenvector of $\Sigma_\xi$ with the largest eigenvalue. In this case, the predictive information objective selects control sequences that spread future states $x_s$ along directions in which there is the largest uncertainty in the estimated scene. It is important to note that we do not need to update $p(\xi \mid y_1^t)$ using observations beyond time $t$ while solving the optimization problem; these observations belong to the same sigma algebra as the one generated from $y_1^t$---they do not contain any new information.
\end{example}

For a system with nonlinear dynamics and more complex observations, e.g., where $y_t$ are images, it is difficult to calculate the optimal control sequence directly. In such cases, we could use reinforcement learning (RL) techniques with predictive information as the return of a trajectory. But this requires training the RL-based controller together with the generative model $p(\yttp \mid \xi)$ and in the same physical scene. In the next section, we will introduce a sampling-based planner that evaluates predictive information across a few control sequences and selects the best sequence.

%% file: method.tex

\section{Methodology}
\label{s:method}

We next describe the three main components of our approach: (a) neural radiance fields, (b) calculating the predictive information, and (c) sampling control sequences for a quadrotor using a differentially-flat output space to maximize the predictive information. We will work in the setting where the quadrotor receives observations that consist of RGB images $y_t$ rendered from the (true) scene from the viewpoint $x_t \in \SE(3)$. In addition, the quadrotor obtains measurements of the ground-truth pixelwise depth and the ground-truth category ($C$ distinct classes) of the object that each pixel belongs to.

\subsection{Neural Radiance Fields (NeRFs)}

\begin{figure}
    \centering
    \includegraphics[width=\linewidth]{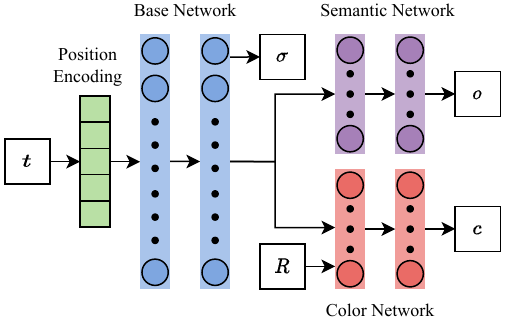}
    \caption{A schematic of the neural architecture used in our approach: in addition to the standard NeRF model that predicts color ($c$) and density ($\s$), we also have an output that predicts categories of the object at location ($t$).}
    \label{fig:semanticNeRF}
\end{figure}

We will represent the scene $\xi: x \mapsto (c, \s, o)$ as a neural radiance field which takes as input a pose $x \equiv (R, t) \in \SE(3)$ and outputs the color $c \in \reals^3$, density $\s \in \reals_+$ and the category $o \in [0,1]^C$ of the object that the point in physical space lies on. The NeRF is parameterized by a multi-layer perceptron (MLP)~\cref{fig:semanticNeRF}. We can think of an input image $y$ as a collection of viewpoints $x \in \SE(3)$ with the same translation $t$ and different orientations corresponding to the plane of image formation. A NeRF encodes the viewpoint using a set of basis functions over the domain, this is called position encoding. While initial implementations of position encoding used Fourier modes~\cite{vaswani2017attention} (i.e., sines and cosines of different frequencies), more recent ones use a multi-resolution hash map of the 3D space encoding~\cite{mueller2022instant}. Our implementation of NeRF for color and density closely resembles the latter. See the original paper for more details.

\paragraph*{Synthesizing new images}
Using a trained NeRF, we can synthesize an image from a new viewpoint $x = (R, t) \in \SE(3)$ using the volume rendering equation as follows. Assume a pinhole model for the camera and imagine a set of rays that emanate from the focus at $t$. A point at a distance $d \in \reals$ from the focus along this ray that has orientation $\delta R R$ can be written as $x(d) = \delta R R\ (d, 0, 0)^\top + t$; we will be interested in rays oriented along $\delta R R$ that densely span the field of view of the camera. The volume rendering equation samples points along this ray while querying the NeRF for color $c(x)$, density $\s(x)$ and category $o(x)$. Transmittance
\beq{
    p(d)= \exp \rbr{-\int_{d(\delta R R)}^d \dd{s} \s(x(s))}
    \label{eq:transmittance}
}
is the probability that a ray travels for a distance $d$ without encountering a solid object, and therefore $p(d) \s(x(d))$ is the probability that the ray stops exactly at $d$. The integration domain begins at $d(\delta R R)$ which is the distance of the image formation plane from the focus along that ray. This pixel has
\beq{
    \aed{
        \text{color}: y_{\text{rgb}} &= \int_{d(\delta R R)}^d \dd{s} p(s) \s(x(s)) c(x(s)),\\
        \text{depth}: y_d &= \int_{d(\delta R R)}^d \dd{s} p(s) \s(x(s)) s, \text{ and }\\
        \text{category}: y_o &= \int_{d(\delta R R)}^d \dd{s} p(s) \s(x(s)) o(x(s)).
    }
    \label{eq:NeRF}
}
NeRFs implement a discretization of these integrals~\cite{mildenhall2020nerf,inplaceICCV2021}. Synthesizing an image from a NeRF is computationally intensive (thousands of MLP queries for many points along may rays for a single image); \cite{mueller2022instant} achieves about 10 fps on a desktop GPU. Techniques like~\cite{li2023nerfacc} accelerate the rendering process by skipping known free space.

\paragraph*{Training a NeRF}
We train the NeRF using a dataset of images, their corresponding viewpoints, dense depth and segmentation masks obtained from a photorealistic simulator of indoor environments. For each image, the training procedure samples rays outwards from the field of view of the camera and queries the MLP for the color, density and category $(c, \s, o)$ at different points on the ray as explained above. The rendered color, depth and category of each pixel are compared to their ground-truth values to calculate
\beq{
    \ell = \l_1 \ell_{\text{rgb}} + \l_2 \ell_{\text{depth}} + \l_3 \ell_{\text{category}},
    \label{eq:NeRF_loss}
}
where we use the $\ell_1$ loss for color and depth, and the cross-entropy loss for the categories. This objective can be used to update the parameters of the MLP using stochastic gradient descent (SGD). Typical NeRF implementations only use the RGB loss but it is useful to include the geometric loss for indoor navigation problems to improve the accuracy of the density $\s(x(s))$ (and thereby free space). We tuned the hyper-parameters $\l_1, \l_2$ and $\l_3$ to ensure that the three terms have approximately equal magnitude during training.

See~\cref{fig:fig3a} for a NeRF trained on a fixed dataset. We also need to update the representation of the scene as new observations are procured. We do so by continuously running SGD using mini-batches where half of the images are sampled from recent observations and the other half are sampled uniformly randomly from past observations~\cite{yu2023NeRFbridge}.

\subsection{Calculating the predictive information}

We calculate a distribution over scenes (NeRFs) using a bootstrap over the collected data to build~\cref{eq:p_xi}. This is a very effective way to build an ensemble of models (we found that $m = [2, 4]$ was sufficient to calculate all quantities accurately). NeRFs are not probabilistic models and therefore they cannot directly provide us with a likelihood of the new observation $p(\yttp \mid \xi)$. They only allow us to obtain images from a new viewpoint, and \emph{only after} the actual observation is received can we calculate~\cref{eq:NeRF_loss} to interpret it as the likelihood of the image. But since the integrals in~\cref{eq:NeRF} are just an expectation, we can refashion them to calculate the variance:
\[
 \textbf{}   \text{var}(y_{\text{rgb}}) = \int_{d(\delta R R)}^d \dd{s} p(s) \s(x(s)) \rbr{c(x(s)) - y_{\text{rgb}}}^2,
\]
with an analogous expression for depth to calculate $\text{var}(y_d)$. Now if we approximate color and depth as having a Gaussian distribution, we can calculate the probability $p(\yttp \mid \xi)$ as well as entropy $\S(\yttp \mid \xi)$ for color and depth easily. The output $y_o$ is already a probability distribution, it is a multinomial distribution over the categories. We also found it beneficial to retrieve an additional observation from the NeRF, namely the occupancy. As the authors in~\cite{sunderhauf2022densityaware} discuss, most rays in the NeRF eventually intersect with some solid surface in a bounded scene so the expected occupancy of every ray, i.e., the transmittance in~\cref{eq:transmittance} should be zero for some large value of $d$. We can model the occupancy $y_{\text{occ}}$ along a ray as a Bernoulli random variable: the ray hits an obstacle eventually with probability $1-p(d_{\text{max}})$ and goes off to infinity with probability $p(d_{\text{max}})$; this way we can calculate the entropy $\S(y_{\text{occ}} \mid \xi)$ as the entropy of the Bernoulli distribution.

\begin{figure*}[t!]
\centering
    \begin{subfigure}[b]{0.55\linewidth}
    \includegraphics[width=0.24\linewidth]{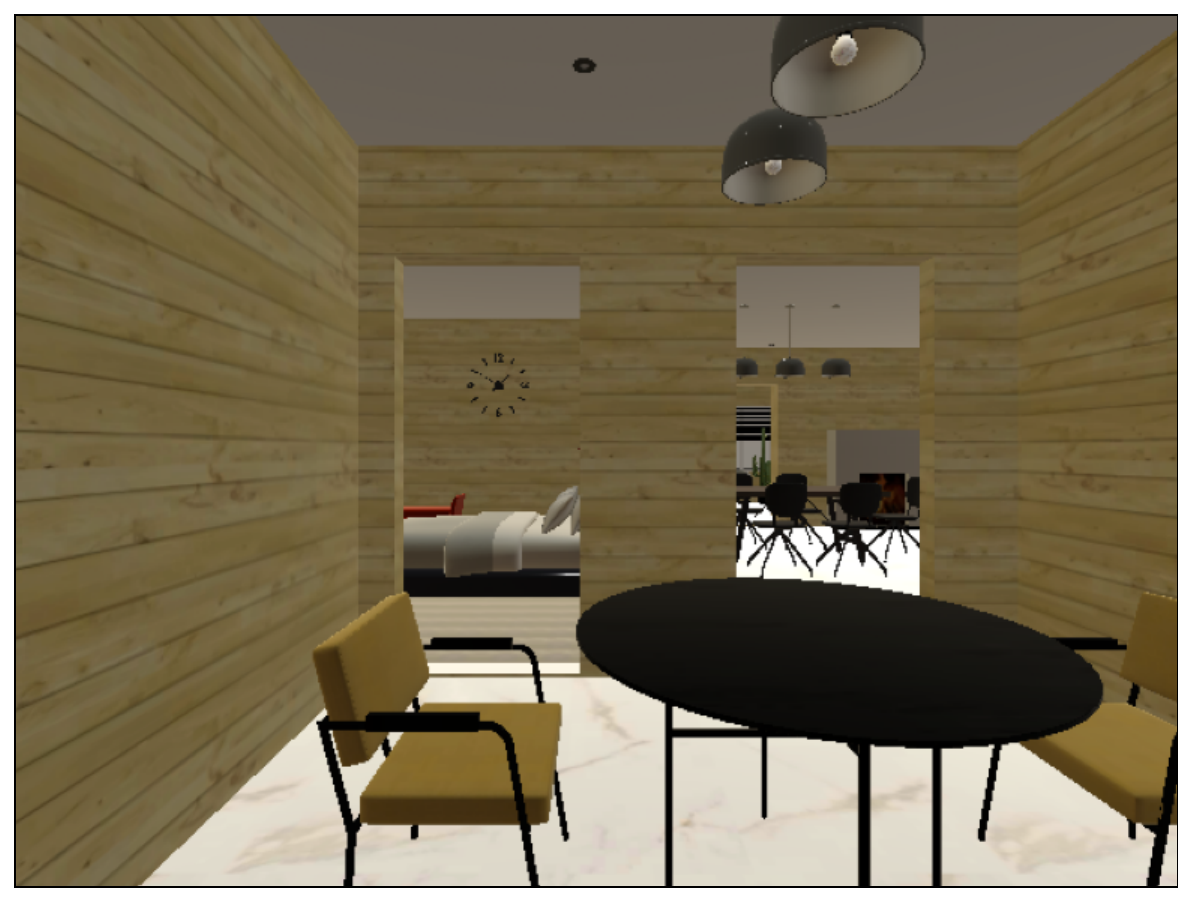}
     \includegraphics[width=0.24\linewidth]{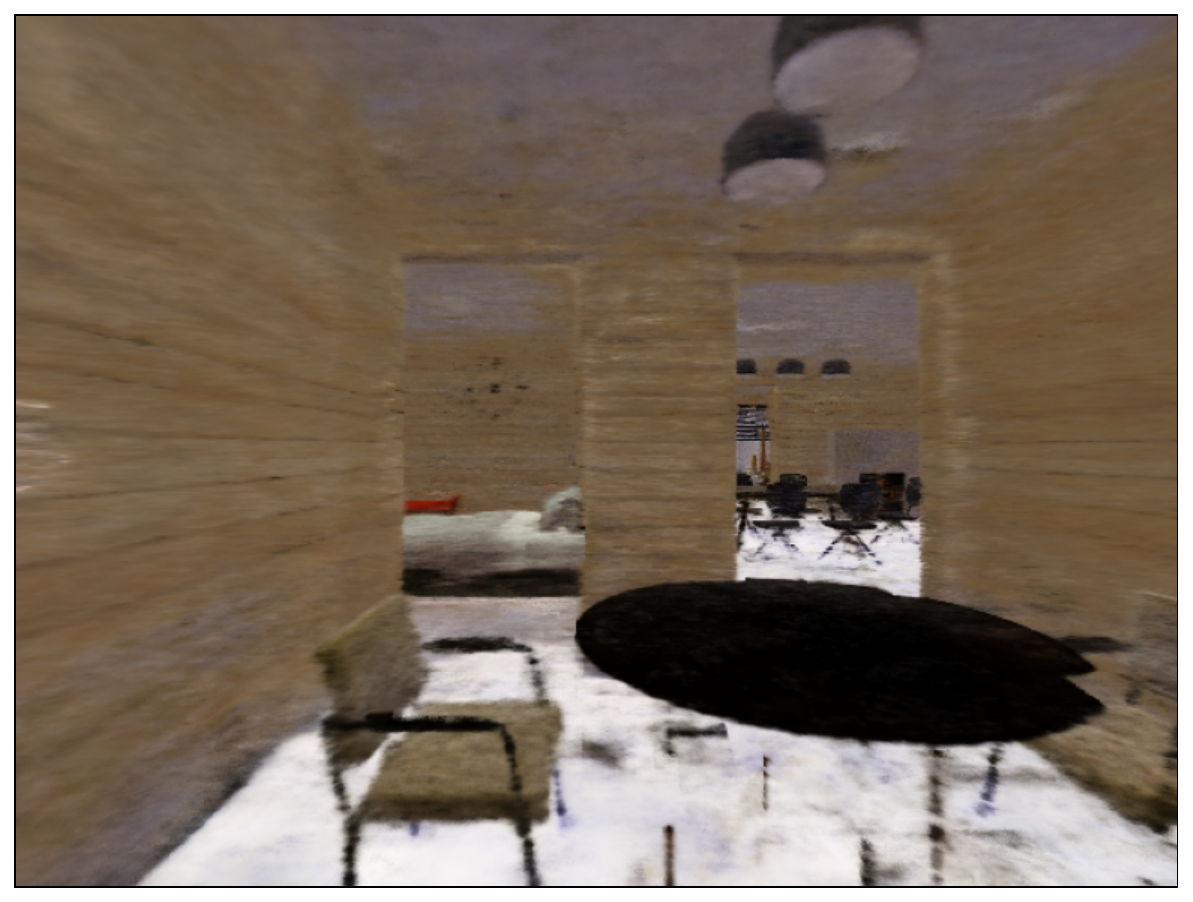}
     \includegraphics[width=0.24\linewidth]{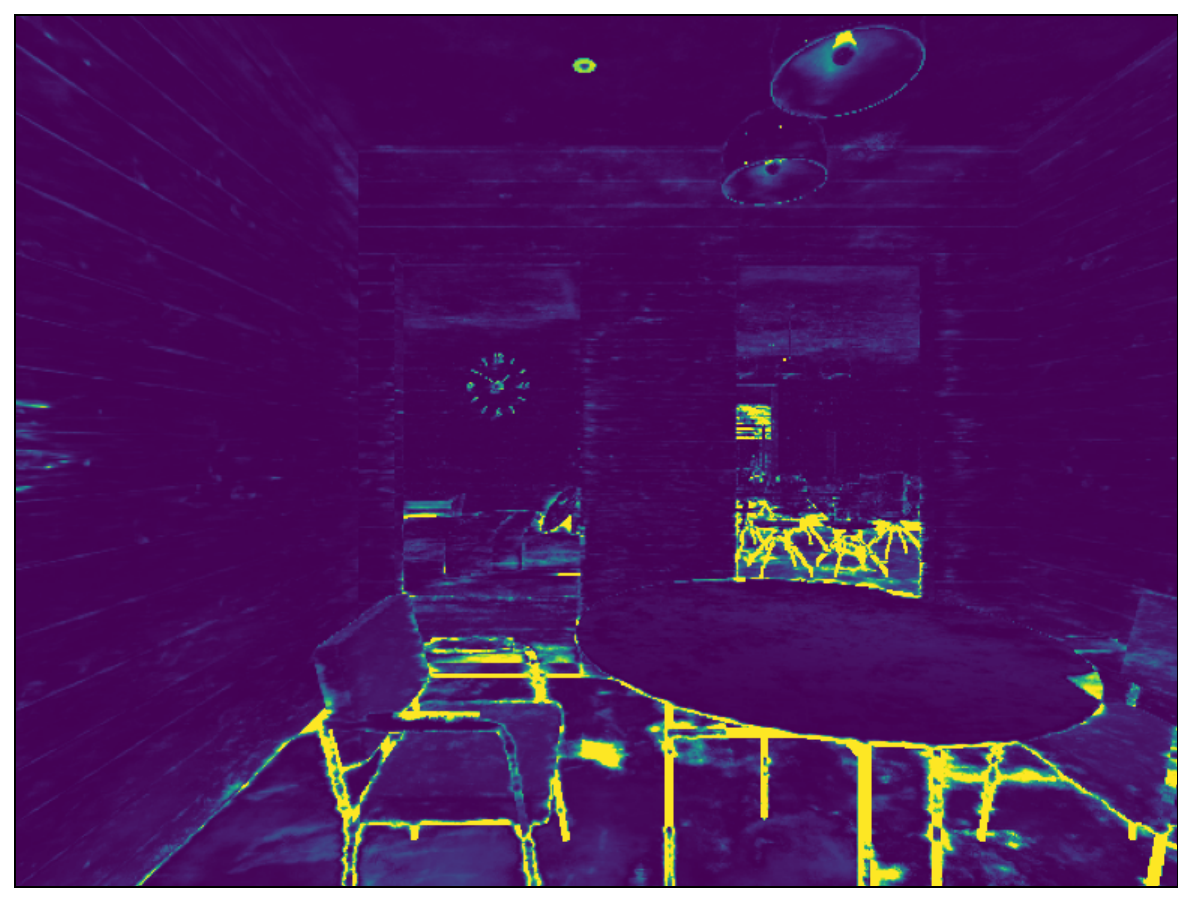}
     \includegraphics[width=0.24\linewidth]{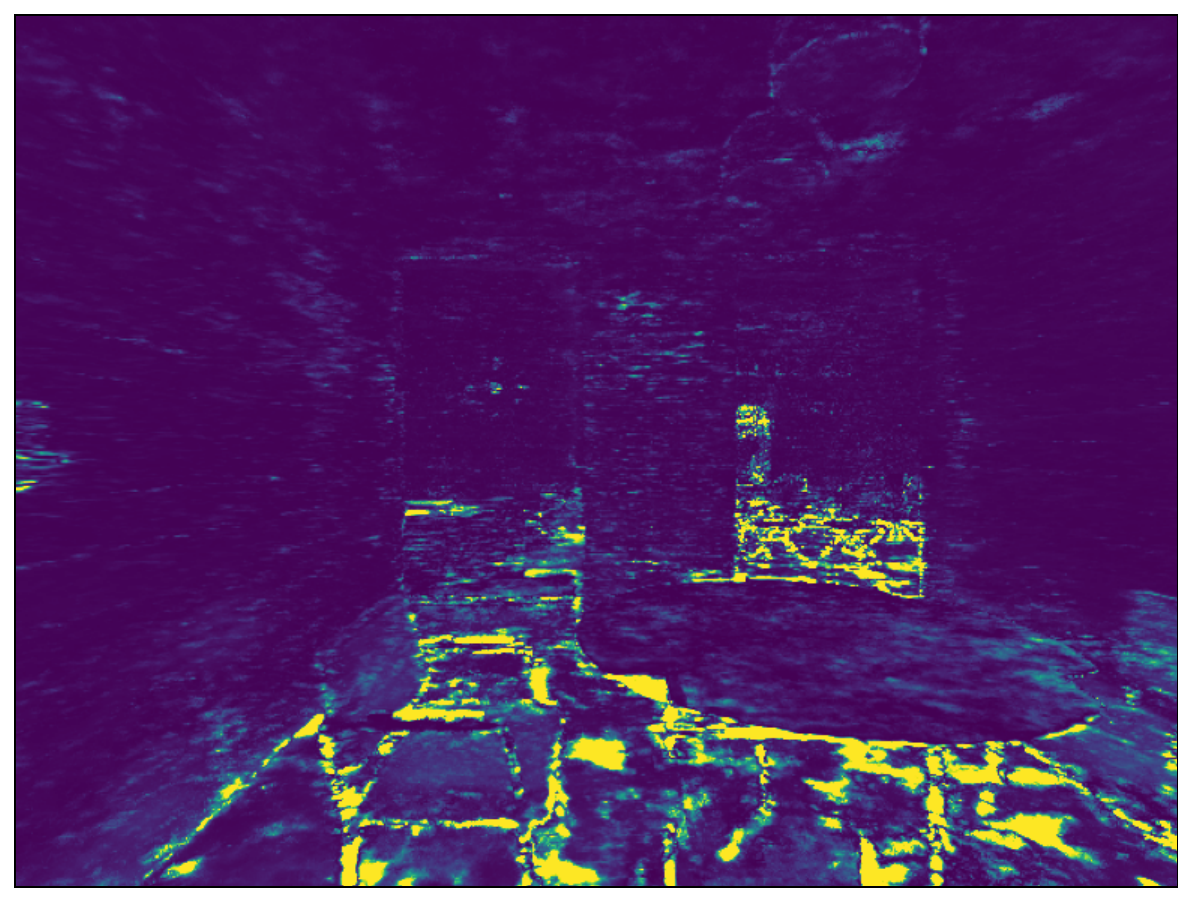}

     \includegraphics[width=0.24\linewidth]{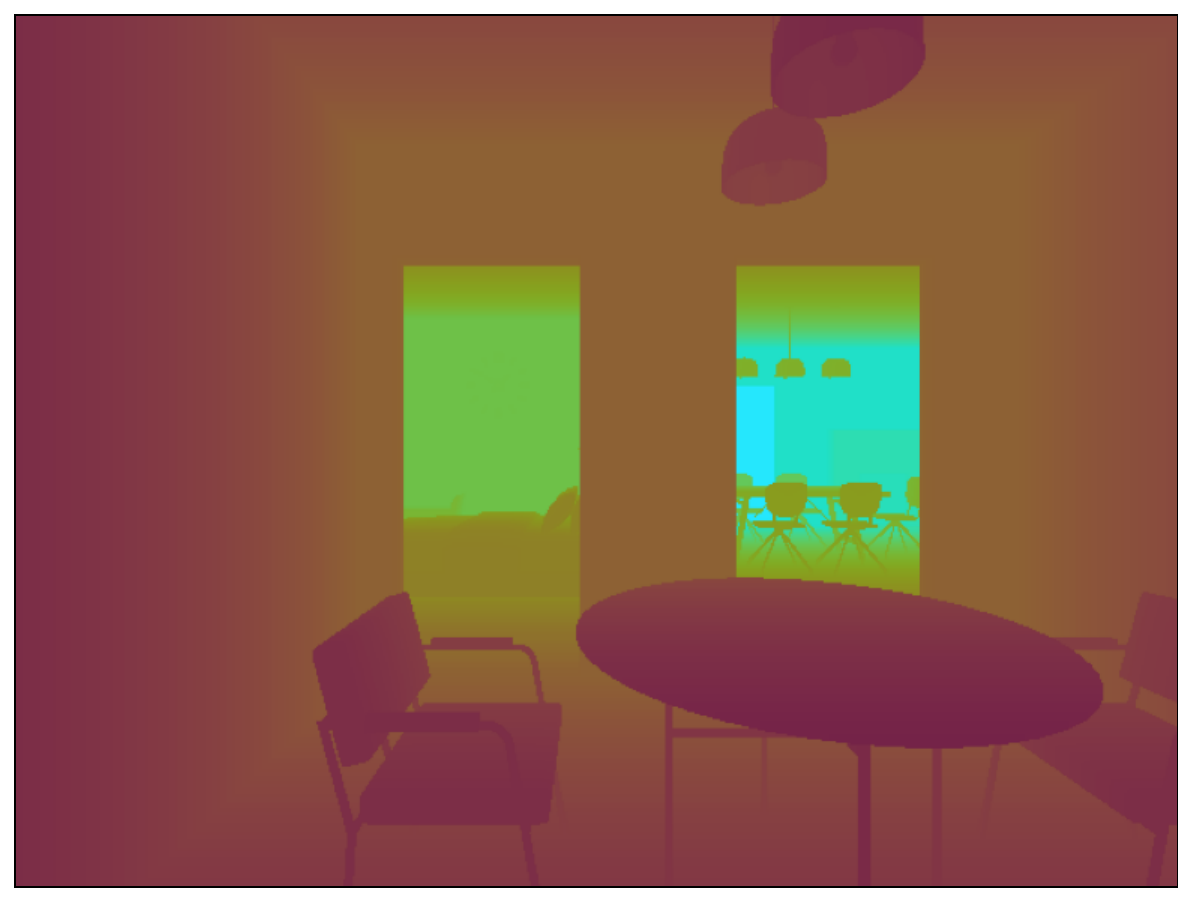}
     \includegraphics[width=0.24\linewidth]{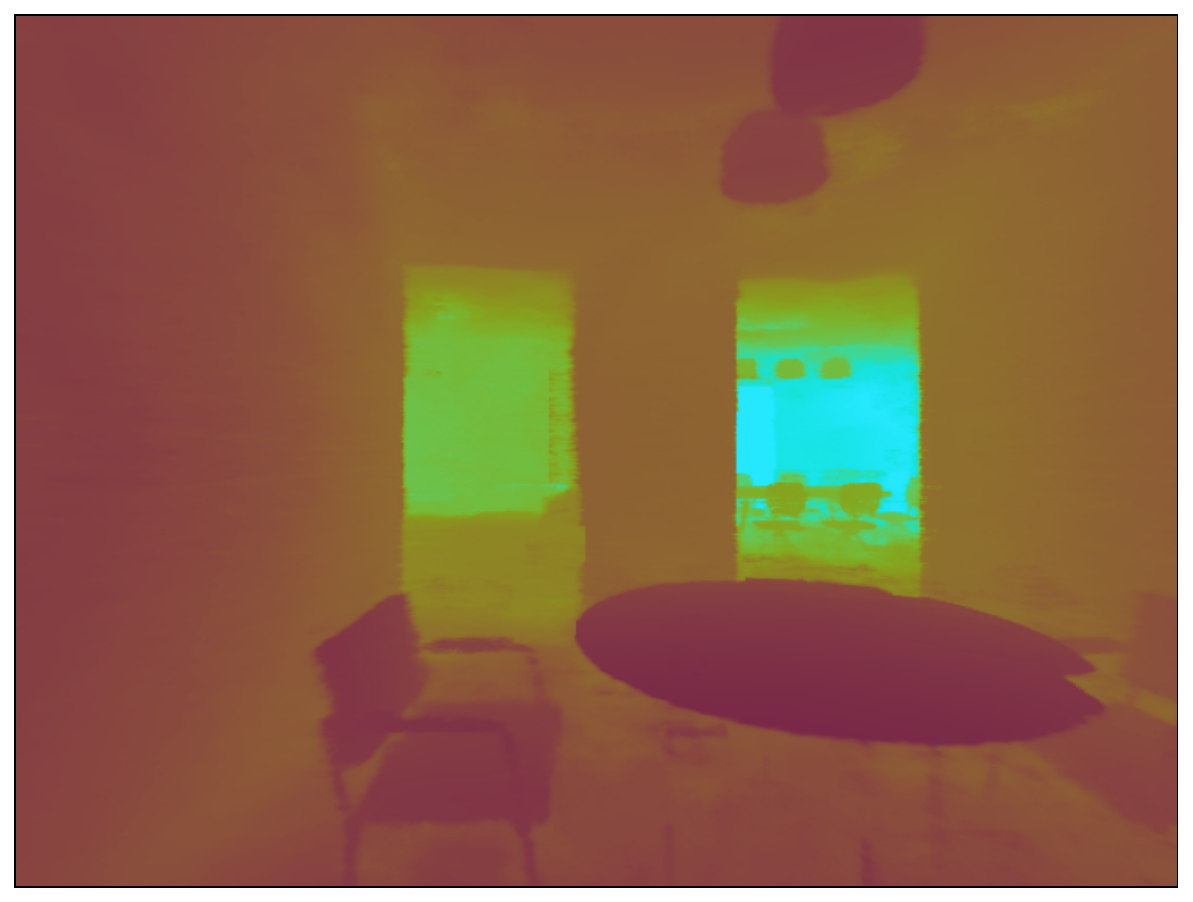}
     \includegraphics[width=0.24\linewidth]{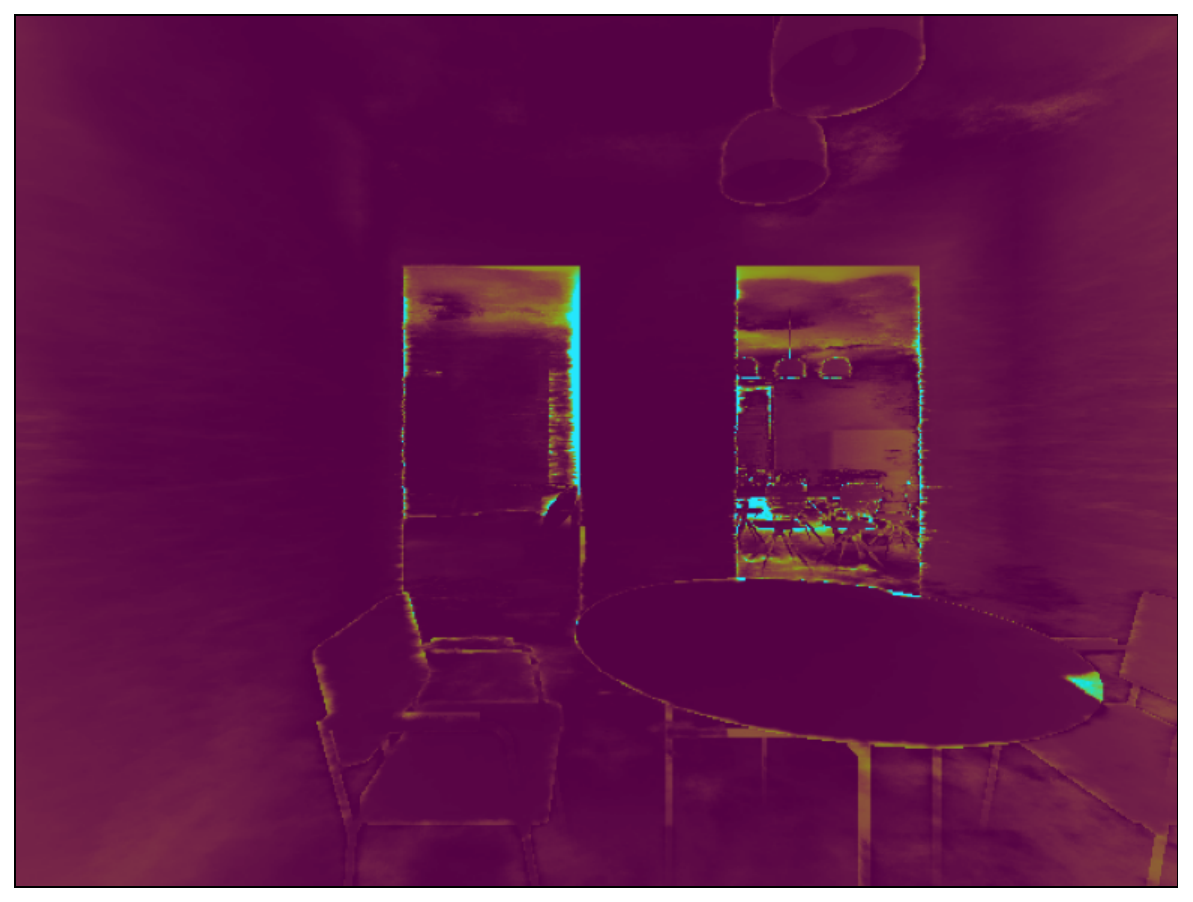}
     \includegraphics[width=0.24\linewidth]{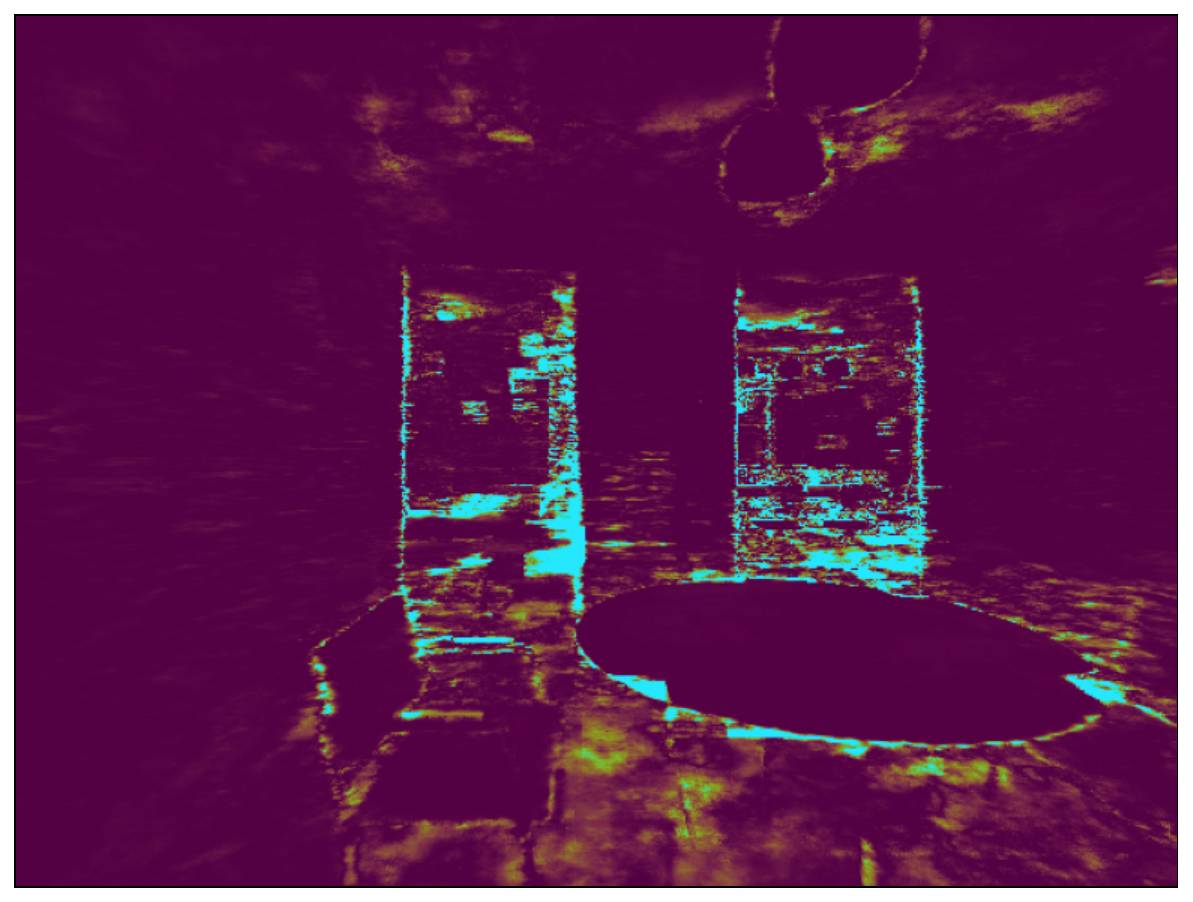}

     \includegraphics[width=0.24\linewidth]{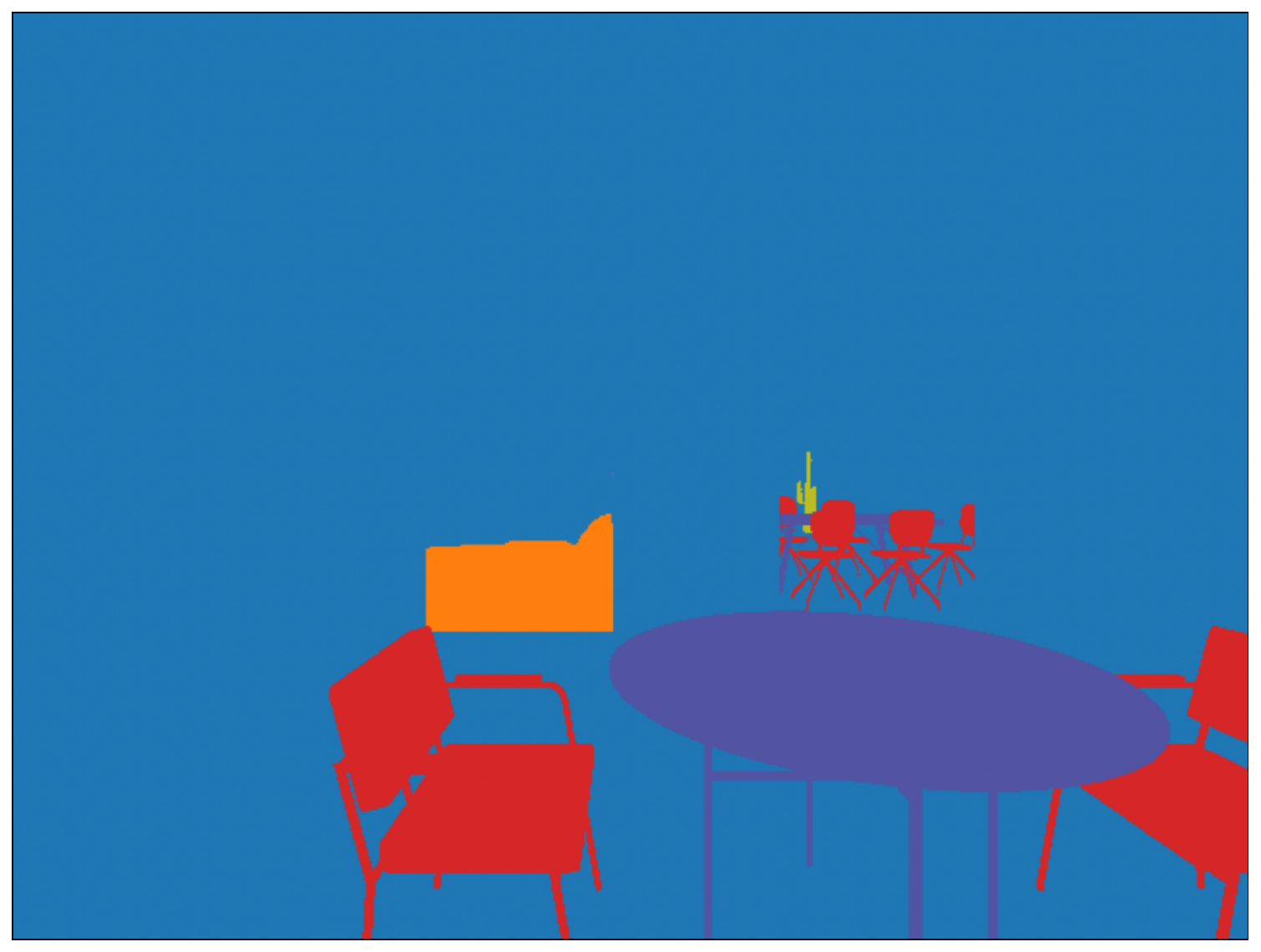}
     \includegraphics[width=0.24\linewidth]{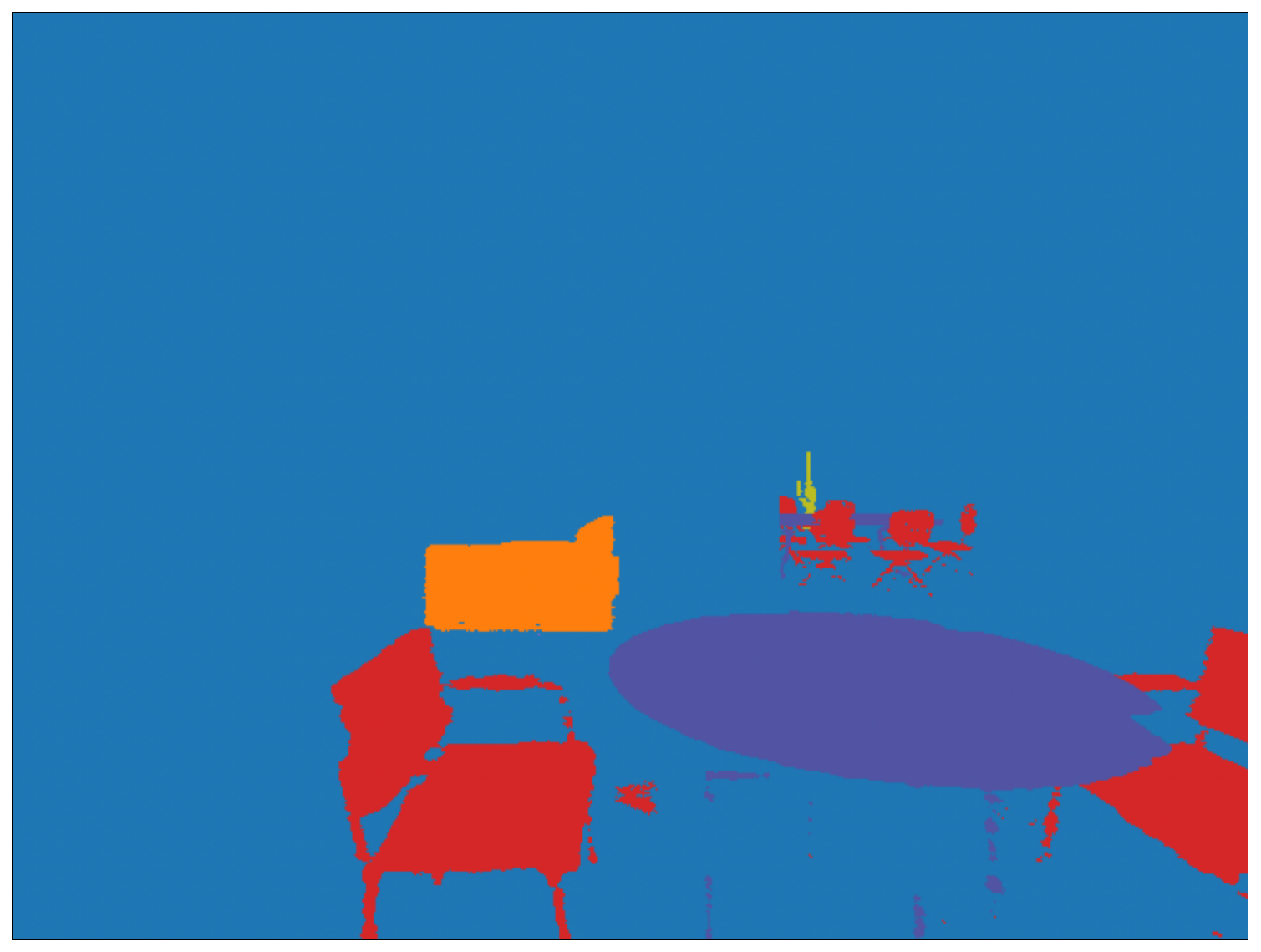}
     \includegraphics[width=0.24\linewidth]{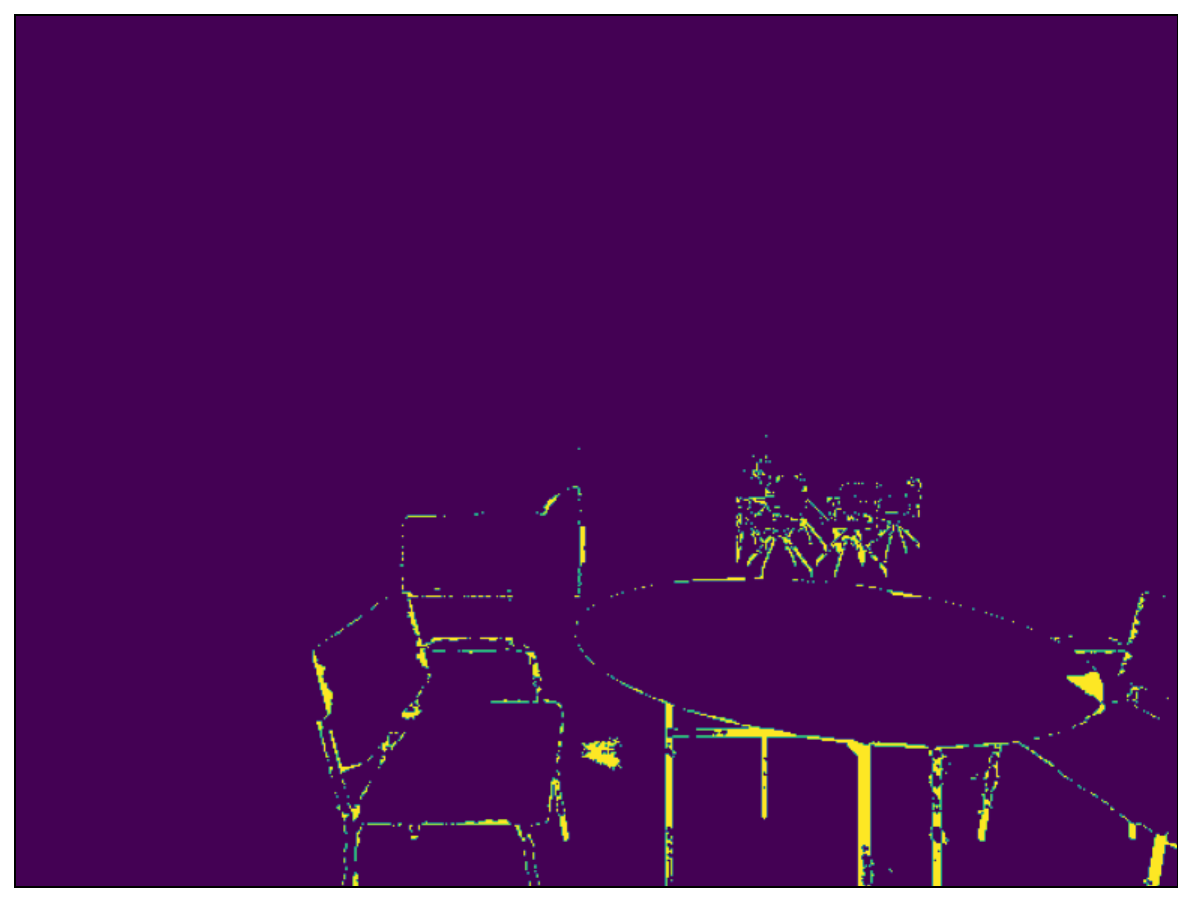}
     \includegraphics[width=0.24\linewidth]{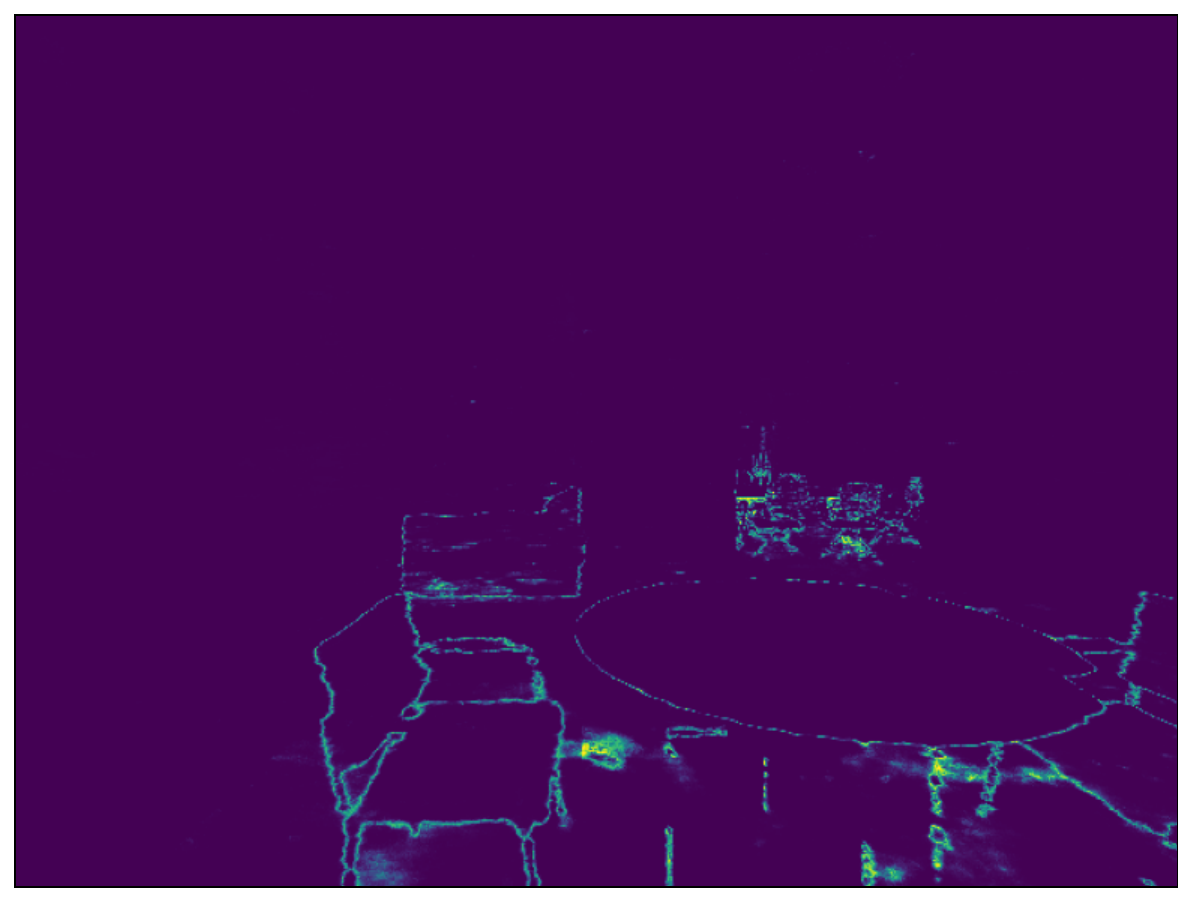}
    \caption{Ground Truth, Synthesized Observation, Error, and Uncertainty}
    \label{fig:fig3a}
    \end{subfigure}
    \begin{subfigure}[b]{0.4\linewidth}
         \centering
         \includegraphics[width=\linewidth]{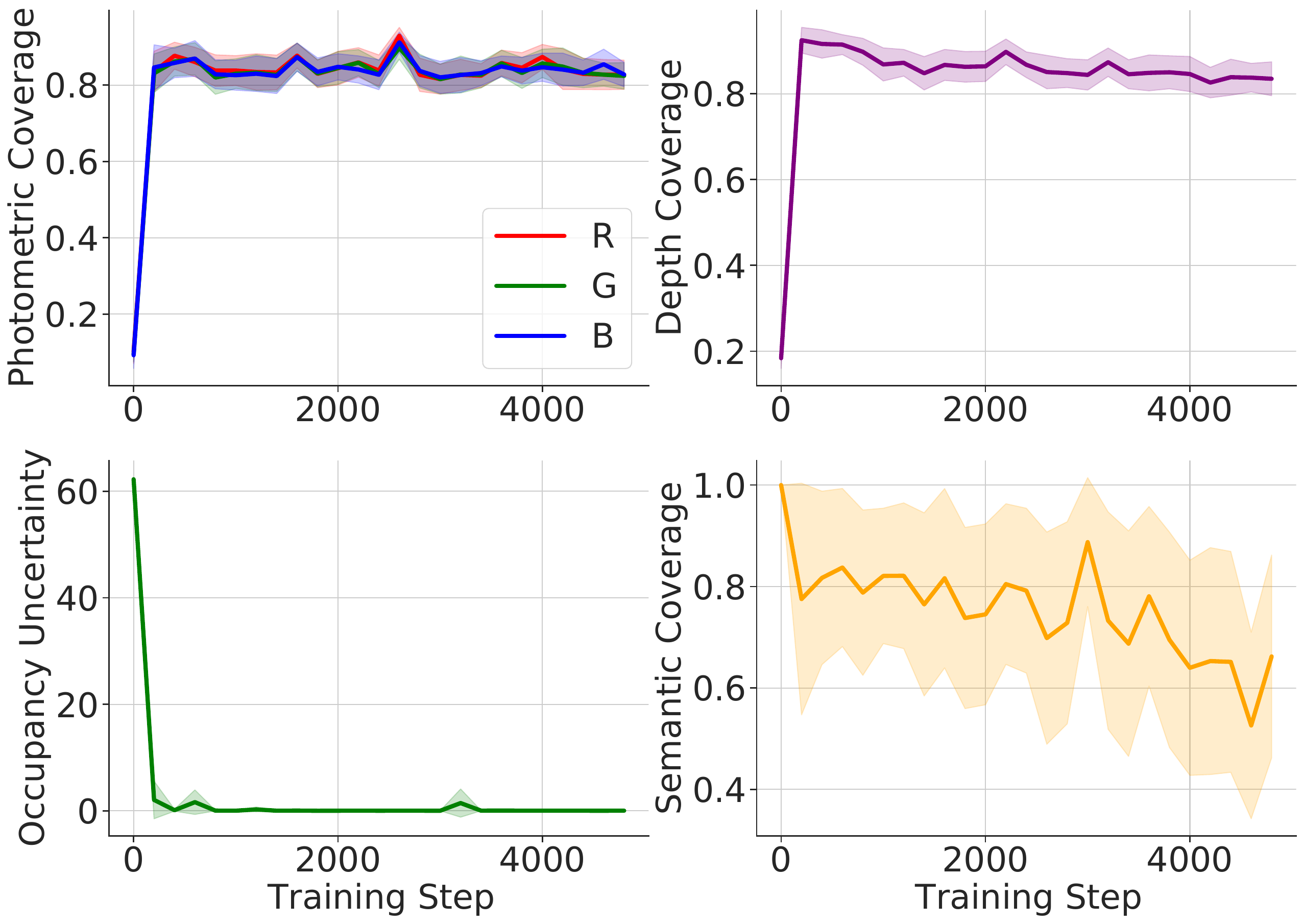}
         \caption{Different kinds of uncertainties}
         \label{fig:depthuq}
     \end{subfigure}
\caption{\textbf{Evaluation of the uncertainty quantification metric.} We train with 39 observations from a Habitat scene (by rotating in place) and test using 18 new viewpoints of the scene. \textbf{(a)} From left to right columns: ground truth observation, NeRF prediction, squared residual, (zero-one loss for categories), and the estimate of uncertainty, for RGB, depth and semantic segmentation (top to bottom row). \textbf{(b)} Coverage of the error bars obtained from uncertainty prediction, i.e., fraction of true RGB values (top left) and depth (top right) that lie within the predicted uncertainty interval, as a function of training steps. Occupancy uncertainty (bottom left) of test observations. Proportion of pixels where the object category was incorrect and uncertain with entropy more than 0.1 (bottom right). Shaded regions denote one standard deviation.
} 
\label{fig:uqe}
\end{figure*}

\paragraph*{Using uncertainty as an error metric}
Using the development of the previous section, we can also calculate the uncertainty of all estimates given by the NeRF. This is very useful as an error metric and allows us to understand the kinds of mistakes that the NeRF makes in its estimates of the RGB intensities $y_{\text{rgb}}$ and the depth $y_d$. In our experiments, we calculate the coverage for color and depth using the one-standard deviation; we evaluate the NeRF based on whether the true quantities lie within one standard deviations $\sqrt{\text{var}(y_{\text{rgb}})}$ and $\sqrt{\text{var}(y_d)}$  of the estimated quantities ($y_{\text{rgb}}$ and $y_d$). The uncertainty over the predicted categories is simply the entropy of the softmax probabilities.~\cref{fig:depthuq} shows the photometric, depth, occupancy, and semantic uncertainties as a function of training steps for a NeRF trained on a set of observations collected by rotating in place. 

\subsection{Sampling controls to maximize predictive information}

We use a standard dynamical model of a quadrotor. At $x \equiv (R, t) \in \SE(3)$, we have
\beq{
    \aed{
    \ddot t &= M^{-1} u^b_{\text{tot}} - g,\\
    \dot R &= R\ \hat \Omega,\\
    \dot \Omega &= J^{-1} \rbr{[u^b_x, u^b_y, u^b_z]^\top - \Omega \cross J \Omega},
    }
    \label{eq:quad_dynamics}
}
where $M$ is the mass, $J$ is the inertia tensor in body frame, $g$ is the gravity vector in global frame, and $u^b_{\text{tot}}$ is the total force by the propellers (in the body frame), and $u^b_x, u^b_y, u^b_z$ are the moments about the axes of the body frame. These dynamics are differentially flat~\cite{MellingerMinSnap} with the four flat outputs being $(t, \text{yaw})$; all other parts of the state and the control inputs can be recovered using algebraic functions of the flat outputs and their derivatives. Flat systems allow us to design trajectories that satisfy initial conditions very easily, e.g., any polynomial that fits the initial and final boundary conditions is a dynamically feasible trajectory (up to control constraints).

Given $N$ waypoints (found using Dijkstra's algorithm in free space) and travel times between them, we formulate an optimization problem that parameterizes each flat output using a 7$^{\text{th}}$ order polynomials and minimizes the integral of the squared snap (fourth derivative of the position) along the trajectory while traveling through all the waypoints. In addition to these waypoint traversal constraints, all derivatives of the flat outputs are constrained to be zero for the initial and final waypoint (quadrotor at standstill). For each intermediate waypoint, we enforce continuity of the first four derivatives of the position $t$ and the first two derivatives of the yaw angle. This gives a quadratic program that can be solved efficiently. We implemented this procedure using rotorpy~\cite{folk2023rotorpy}.

\begin{figure}
\centering
\includegraphics[width=0.32\linewidth]{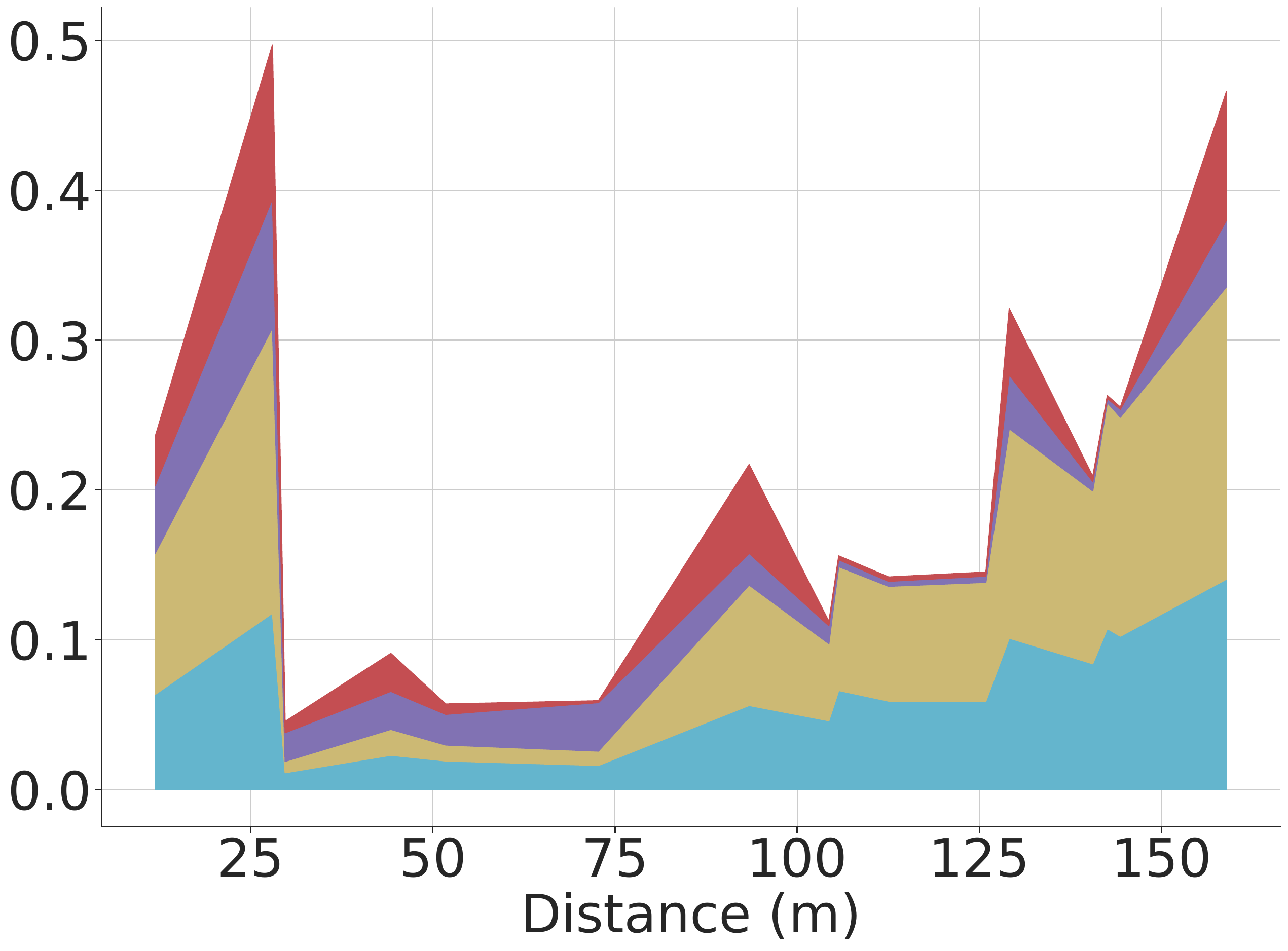}
\includegraphics[width=0.32\linewidth]{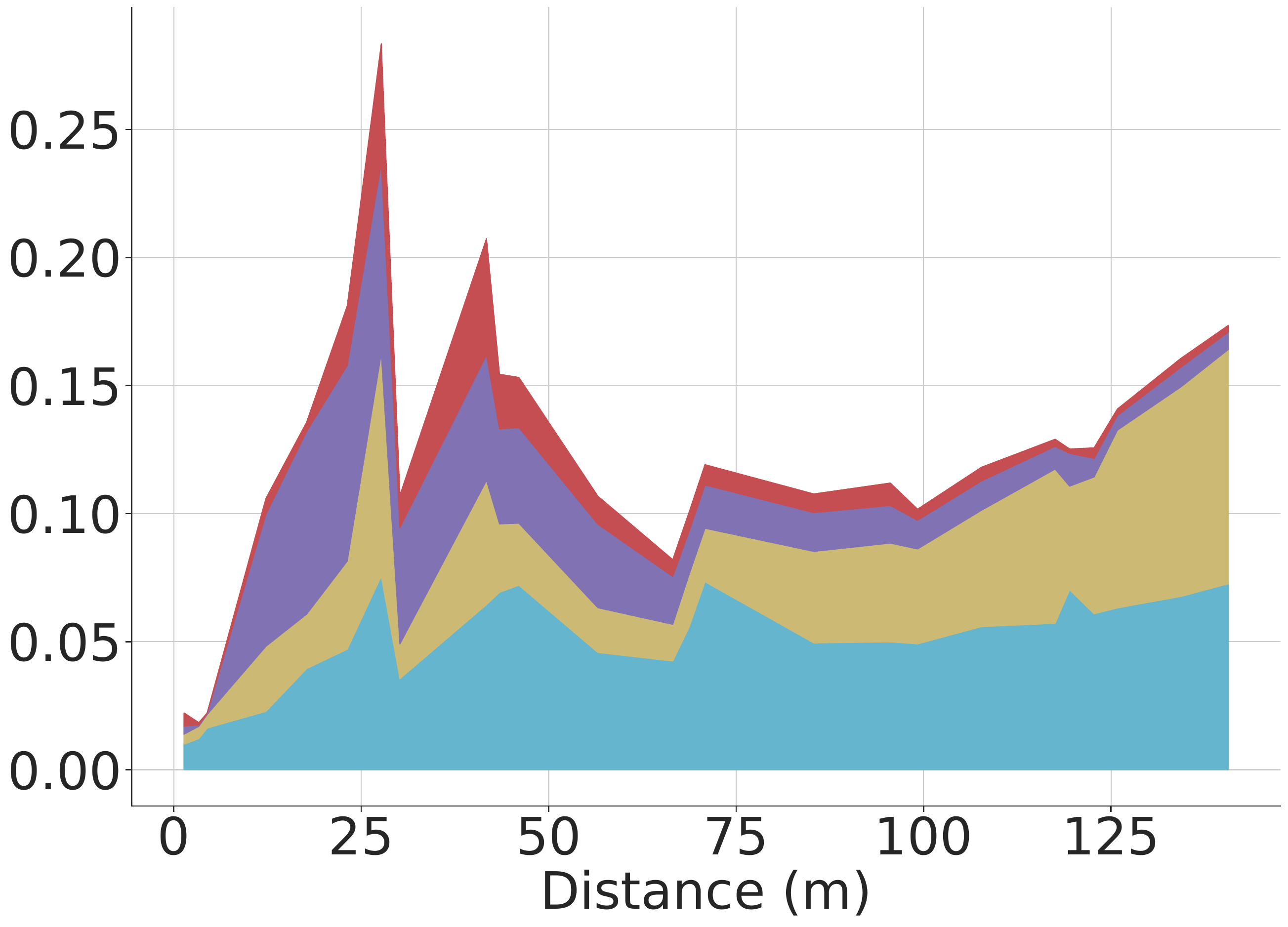}
\includegraphics[width=0.32\linewidth]{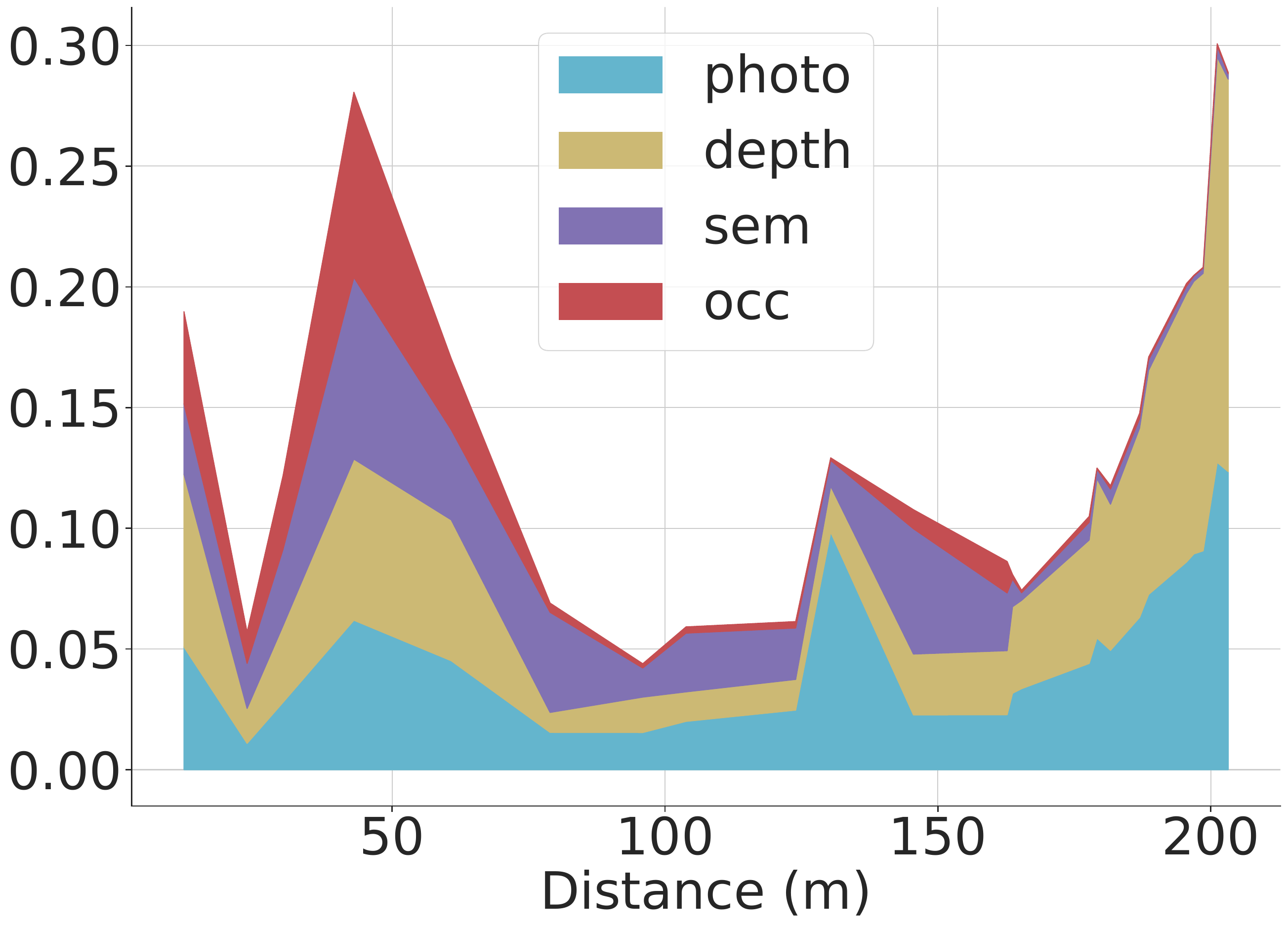}
\vspace*{0.25em}

\includegraphics[width=0.32\linewidth]{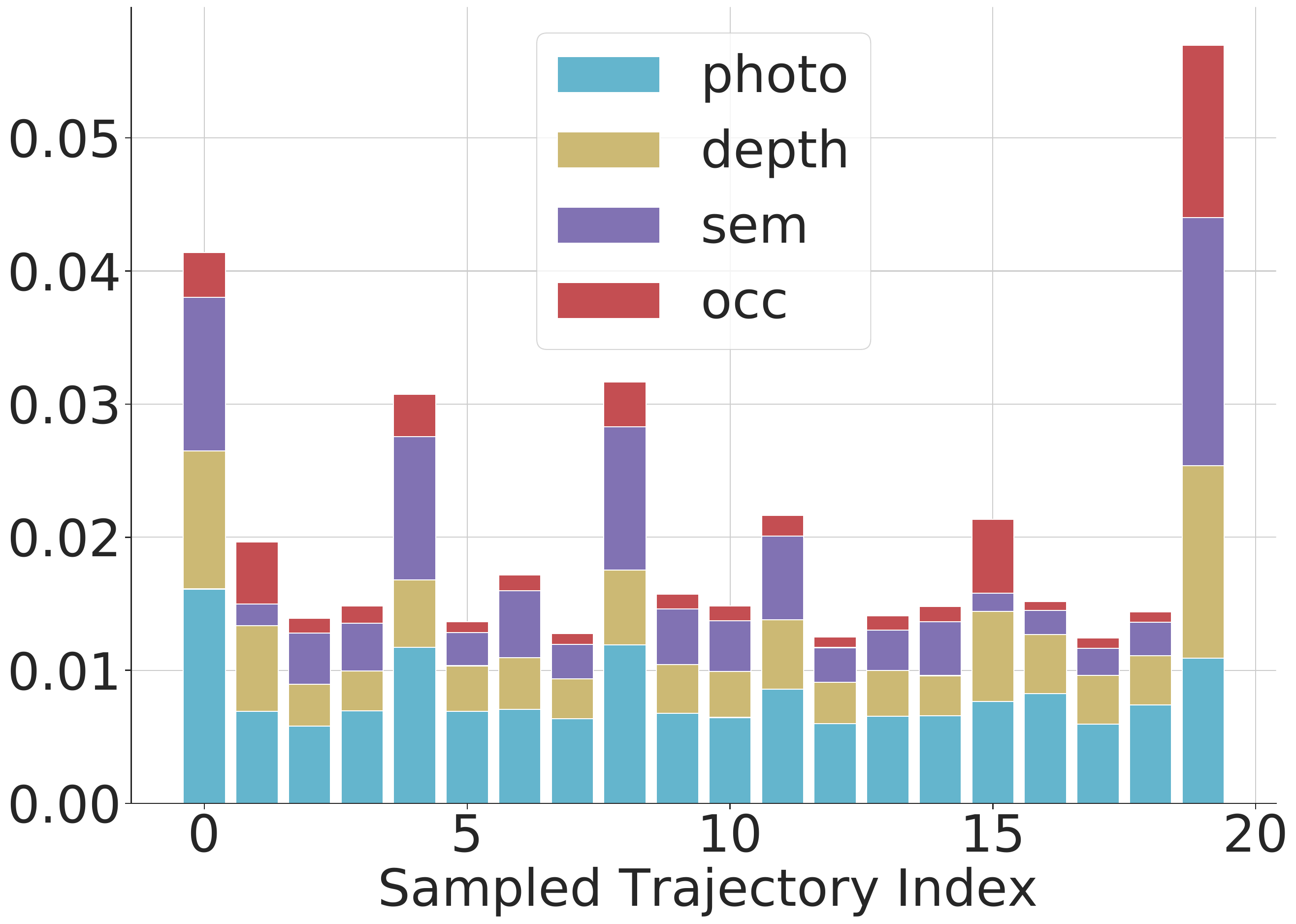}
\includegraphics[width=0.32\linewidth]{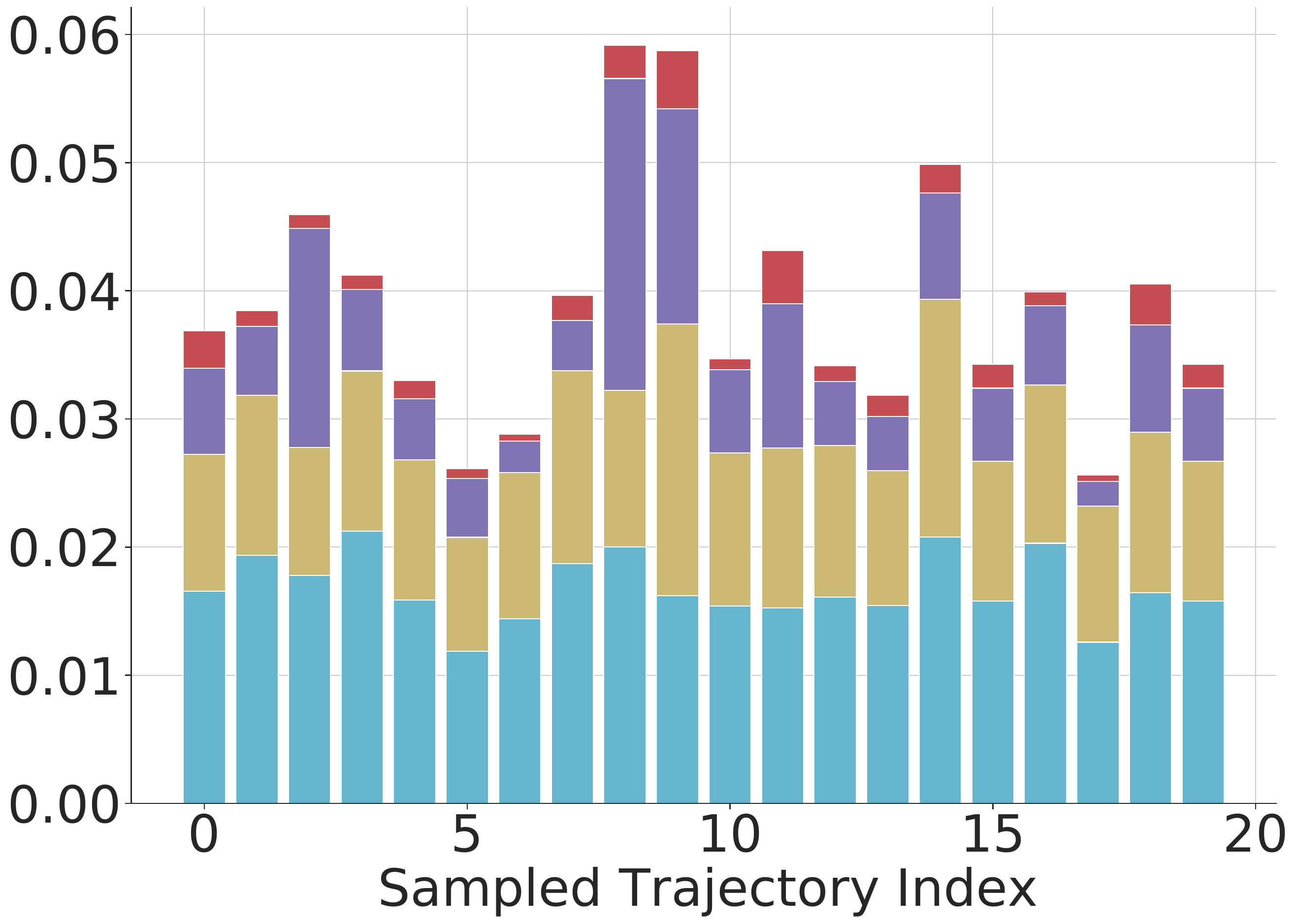}
\includegraphics[width=0.32\linewidth]{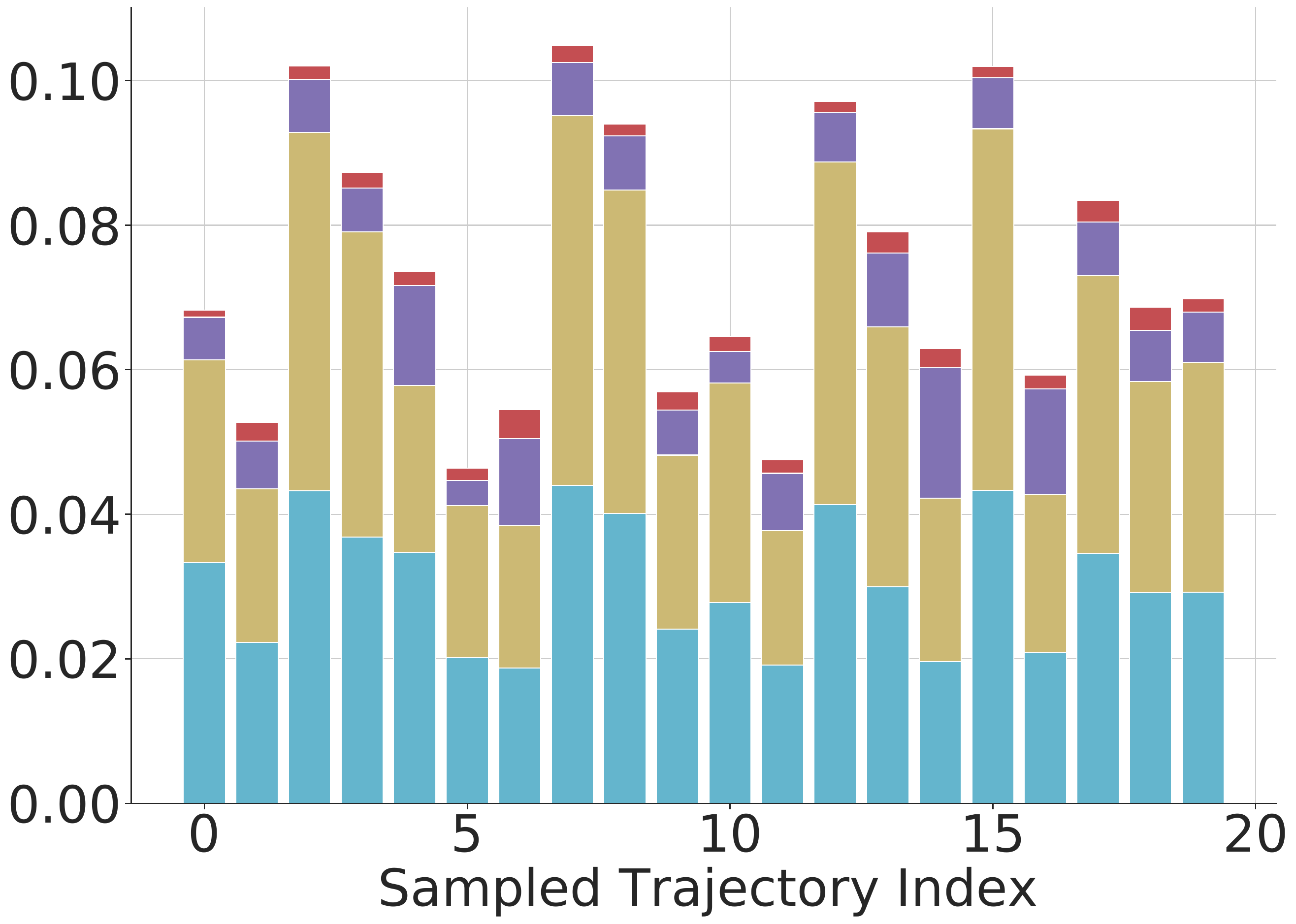}
\vspace*{0.25em}

\includegraphics[width=0.32\linewidth]{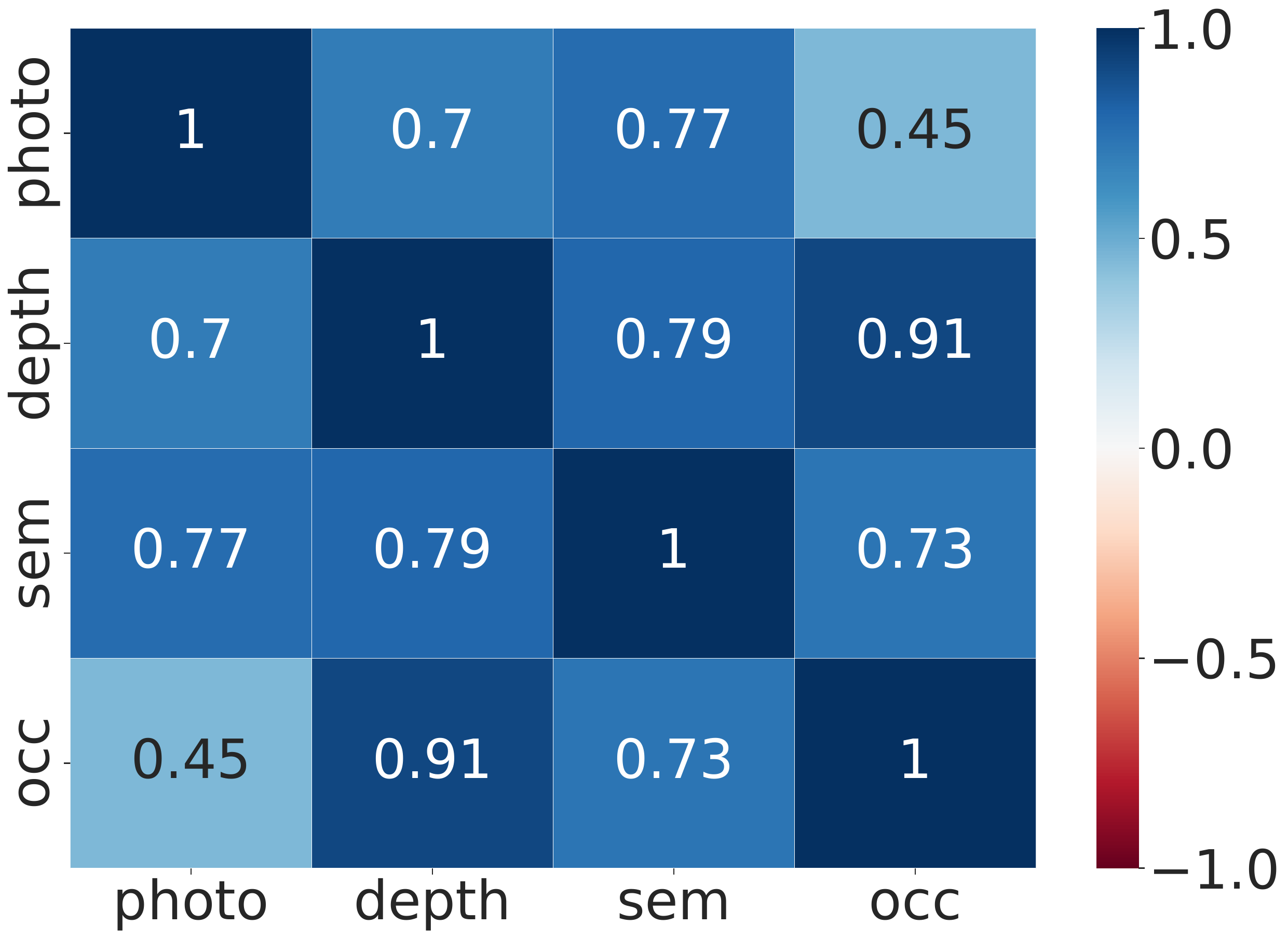}
\includegraphics[width=0.32\linewidth]{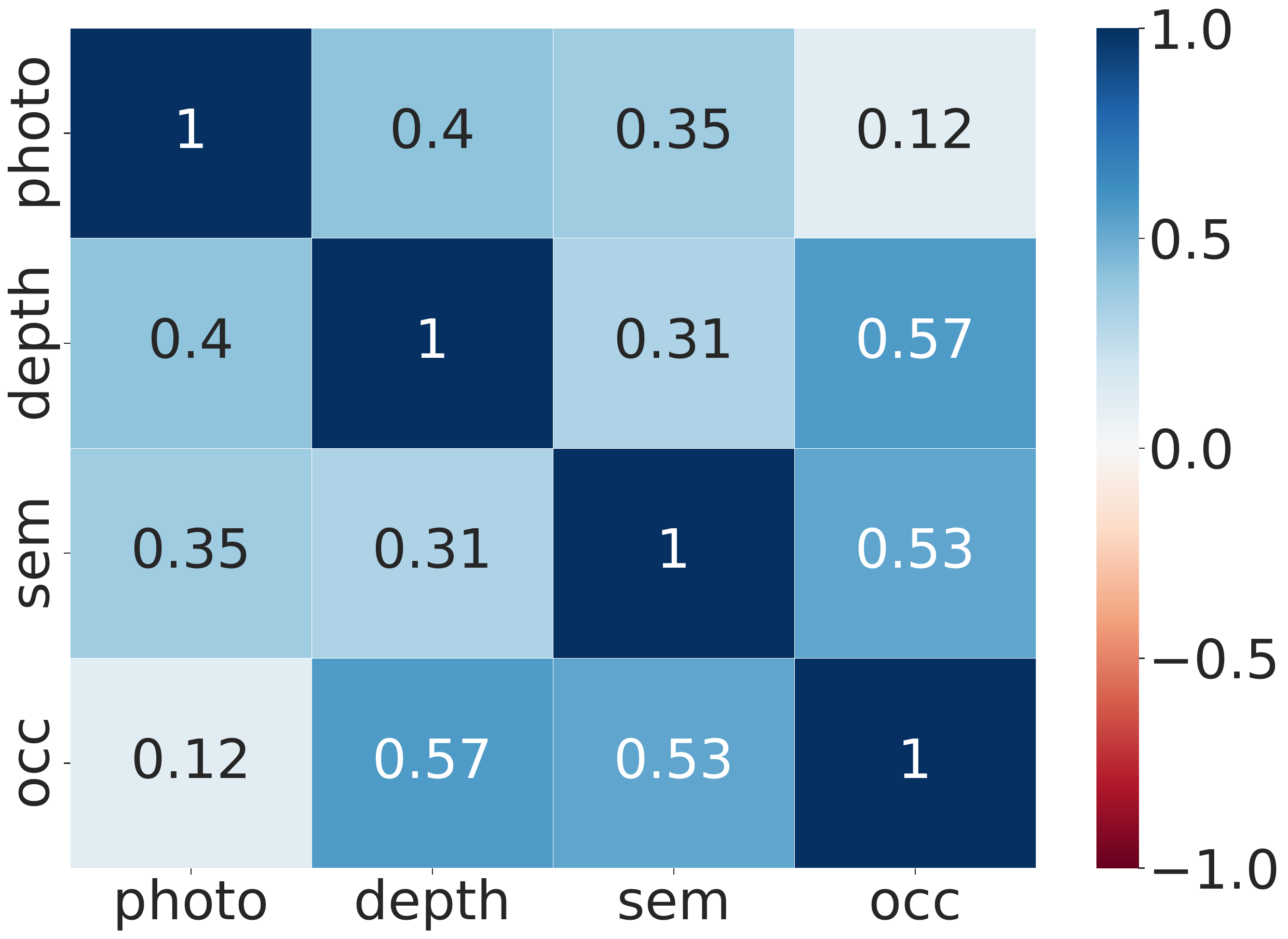}
\includegraphics[width=0.32\linewidth]{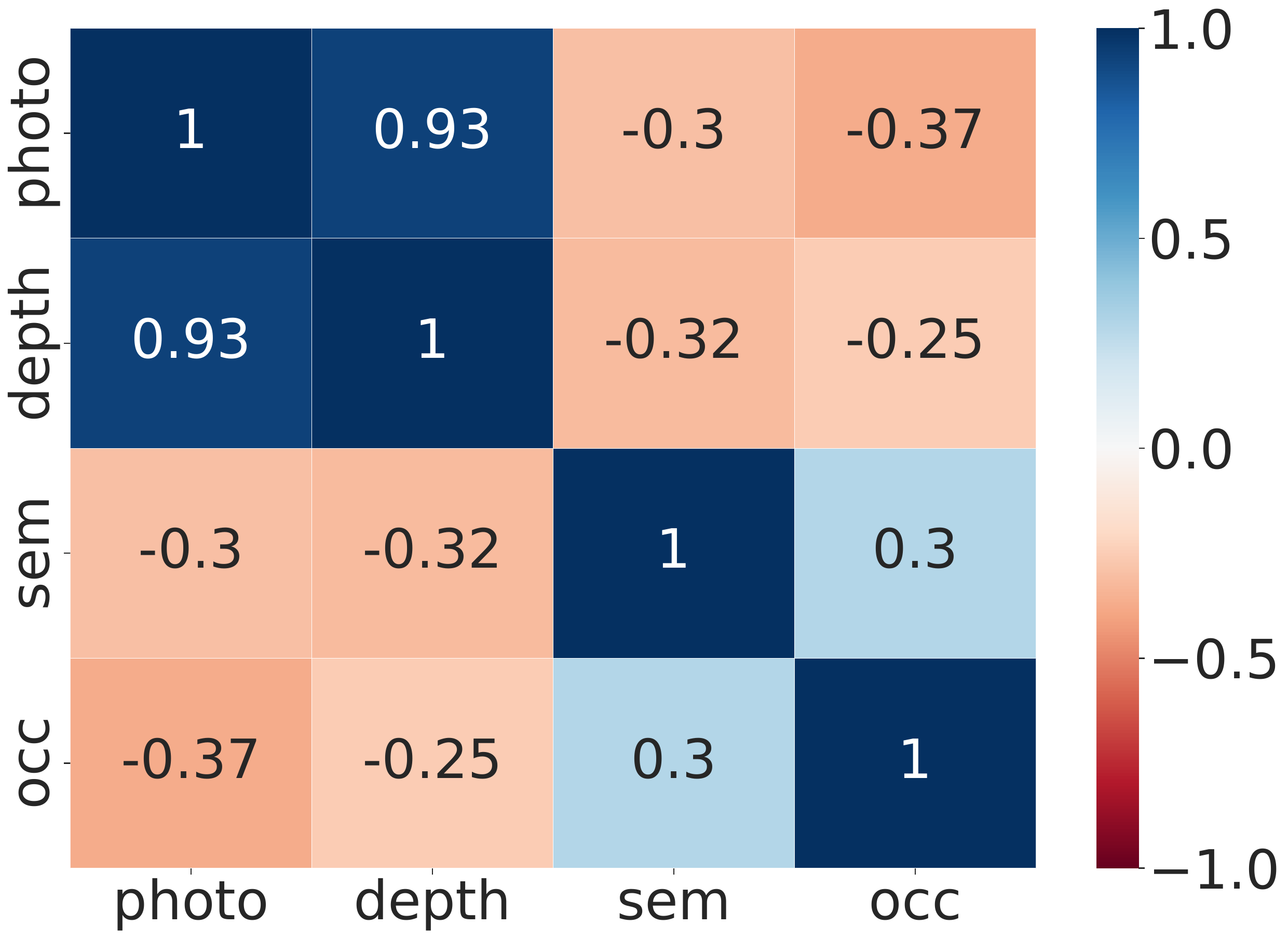}
\vspace*{0.25em}

\caption{
Top: Predictive information ($\Ipred$) of the executed trajectories for the three different scenes.
Middle: $\Ipred$ integrated along trajectories at the beginning, middle, and end of active perception for scene 3 (top right).
Bottom: Correlation among the four types of $\Ipred$ at beginning, middle and end for scene 3 (top right).
}
\label{fig:uncertaintybarchart}
\end{figure}

\paragraph*{Moving like a ballerina to maximize visual information}

We build $N=20$ dynamically-feasible trajectories $(x^{(i)}(t))_{t=0}^T$ by choosing final waypoints uniformly randomly over the free voxels of the NeRF. For each trajectory, we found it useful to enforce constraints upon the yaw angle whereupon the quadrotor rotates by an angle $2 \pi$ along the trajectory (this gives us images with non-overlapping field of views). In addition to this we perform another rotation of $2 \pi$ around the end point of each trajectory. These trajectories are similar to the motion of a ballerina who rotates on her way to the end point. For each trajectory we sample 20 observations from the NeRF to obtain the future observations $\yttp$, calculate the predictive information~\cref{eq:pred_i} of color, depth and semantics, to pick the trajectory with the largest $\I_{\text{pred}}(\yttp, \yt)$. To control the relative importance of the different observation modalities in the active perception objective, we weigh the predictive information of each modality with manually tuned weights; we found $\I_{\text{pred}}=\I_{\text{photometric}}+\I_{\text{depth}}+3\I_{\text{semantics}}+2\I_{\text{occupancy}}$ to provide good results. In~\cref{fig:uncertaintybarchart}, executed trajectories initially have relatively higher occupancy and semantic predictive information which implies a phase of quickly exploring unseen regions. During later steps, the agent is revisiting areas where photometric predictive information are more dominant; it implies a phase of forming a more detailed understanding of visited regions. The correlations among types of uncertainties also vary given different planning steps and locations. Each uncertainty performs a different role in different stages of active perception. At the beginning, the correlation is high since unexplored regions should have high predictive information and explored region should have low predictive information. In later steps, occupancy and semantic labels are learned well, therefore we get less predictive information, however, photometric and depth predictive information increases as we learn more fine-grained color and geometry.

\subsection{Implementation details}

Each experiment begins with the quadrotor spawning at a random location in the environment and collecting an initial set of observations by performing a $2 \pi$ yaw rotation as seen in~\cref{fig:uqe}. In the initialization phase, the NeRF is trained for 4,000 steps/weight updates using this initial set to build a rudimentary map of the local neighborhood, after which the exploration phase begins. During each planning iteration, the quadrotor calculates the predictive information over 20 sampled trajectories with corresponding synthesized images, picks the best trajectory among them, and executes it. During the trajectory it records the actual observations and adds them to the dataset. The NeRF is trained for 2,000 steps after each trajectory (as we discussed, around 50\% of steps are trained on these recently collected observations). As mentioned in~\cref{predinfo}, the predictive information grows as more and more details of the scene are learned. The experiment concludes when the predictive information of the past five iterations all exceed a user-chosen threshold. We found 0.1 to 0.2 to be a good range for threshold. We train the NeRF for a further 20,000 steps on all data at the end to improve the map.

We use only one level of the multi-resolution hash table to encode location $x \in \mathbb{R}^3$. Each NeRF in the ensemble uses a two-layer MLP with 128 neurons per layer for predicting the density $\s(x(s))$, 64 neurons per layer each for color $c(x(s))$ and semantic segmentation $o(x(s))$. We maintain a voxel grid (voxel size 0.2 m) computed using NeRF density output to accelerate ray marching~\cite{li2023nerfacc} and sampling free space for trajectory planning. We noticed that larger NeRF models fit occupied space well but they typically presume that unexplored and unseen parts of the 3D space are free space (i.e., small $\s(x(s))$ presumably due to initialization) while smaller models do not fit occupied space accurately but also do not over-estimate free space. An accurate understanding of free-space is critical for motion planning, even if it is not often the focus in computer vision. To counteract this, we build the ensemble in~\cref{eq:p_xi} with one large NeRF and one small NeRF with half the number of hidden neurons. The free space is taken to be the intersection of free space in the voxel grids of the two models. We further dilate obstacles by one voxel to avoid collisions during planning. We use a 2D occupancy grid at a height of 1.5 m above the ground to compute the shortest path using Dijkstra's algorithm\cite{sakai2018pythonrobotics}. After obtaining the path, we assign time to the path with a constant acceleration of +1 m/s$^2$ and -1 m/s$^2$ at the beginning and the end, and a constant velocity of 0.5 m/s for the rest of the trajectory. We can now obtain a dynamically feasible trajectory for the quadrotor using the procedure described previously. We did not implement a controller to track the trajectories; but it has been shown such trajectories can be tracked well~\cite{MellingerMinSnap}.

%% file: experiments.tex

\section{Simulation Experiments}
\label{s:experiments}

We use a physics-enabled simulator named Habitat~\cite{szot2021habitat} to simulate a quadrotor that navigates realistic 3D indoor environments from the Habitat Synthetic Scene Dataset~\cite{khanna2023hssd}. We present results on three different scenes (scene 1 id: 102344250; scene 2 id: 102344529; scene 3 id: 102344280). For each observation, in addition to the RGB image, we assume that the quadrotor gets access to the ground-truth pose, depth map and semantic masks denoting objects of different types from the simulator.

\subsection{Object localization}
This task involves classifying and localizing objects in the scene and it will be used to evaluate different exploration strategies. This task requires both comprehensive coverage of the scene to discover as many objects as possible while still obtaining sufficient observations of each object to achieve accurate object localization. A 3D semantic voxel map is built from the depth and semantic observations collected during exploration. We associate voxels to individual objects using density-based spatial clustering to obtain the estimated location of objects and their corresponding semantic labels. The estimated locations of the identified objects are compared against the ground-truth locations from the simulator for evaluation. A good exploration procedure should accurately locate all objects in the scene while traveling the shortest distance possible. The number of objects correctly localized ($<$ 0.5 m within ground-truth) as a function of the distance traveled by the quadrotor is a figure of merit for this task.

We compare our approach with two baselines. First, the \emph{Frequency} method samples new locations based on the number of times the field of view the camera saw this voxel in the past~\cite{kopanas2023improving}. A 2D occupancy grid at 1.5 m height is used to keep track of this information for voxel $c(i)$ and the final waypoint of each trajectory is chosen with a probability that is proportional to $\exp(- \beta (c(i) - \min_{j \text{ is free}} c(j)))$, with $\b$ being large if we want to reach seldom observed voxels with a higher probability. This strategy assumes that information gain of a voxel decreases monotonically with its observation frequency; it also does not distinguish between voxels with different photometric, or geometric information.
Second, the \emph{Frontier} method uses the popular frontier-based exploration strategy~\cite{yamauchi1997frontier}. This method uses a 2D occupancy grid to define the boundary between free and unexplored voxels. At each iteration, it detects frontiers that indicate unexplored areas to go to. Frontiers are recalculated using the updated map after each step. This strategy ensures that the robot constantly moves towards unexplored regions, and eventually visits all accessible parts of the scene. In our implementation, the quadrotor plans trajectories between frontiers, collects observations along each trajectory and also performs a $2 \pi$ yaw rotation at the end of each trajectory. The closest accessible and unvisited frontier is chosen at each step\footnote{We did not do so, but we can implement a frontier exploration strategy using the NeRF by noting that the occupancy uncertainty along rays from a viewpoint is high if and only if the camera points to an unexplored region. Therefore, if we select the final waypoint to have maximal occupancy uncertainty, this will be akin to frontier-based exploration.}.

\subsection{Scene reconstruction}
\begin{figure}
\centering
\includegraphics[width=0.53\linewidth]{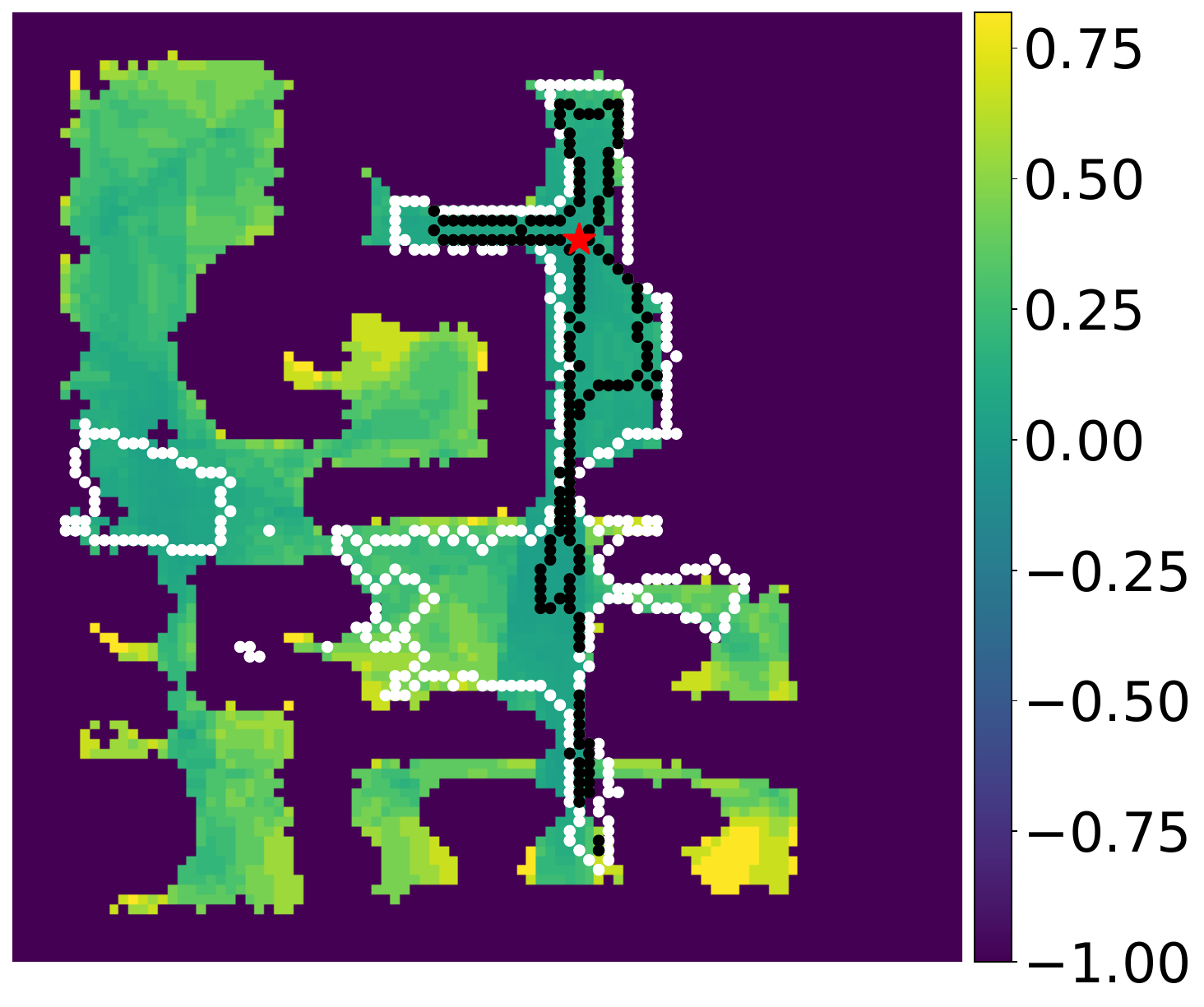}
\includegraphics[width=0.44\linewidth]{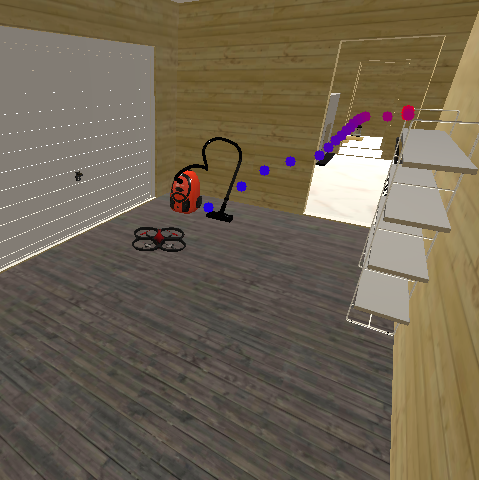}
\caption{\textbf{Exploring doorways while performing active mapping:}
Cross section of the Instant-NGP voxel grid at 1.5 m height after exploration is shown on the left. The quadrotor starts at the red star. Blue pixels with value -1 are obstacles and the rest are free space. The color of a pixel is proportional to the number of times the field of view of the camera observes this pixel. This observation frequency is used in the frequency baseline. The black boundary in the left image denotes the learned map at initialization (without the quadrotor moving, using a few viewpoints). Right after the first trajectory is executed (seen in the right image), the quadrotor discovers a lot of free space in the scene, this is the region enclosed in the white boundary.}
\label{fig:scene3map}
\end{figure}

\begin{figure}
    \centering 
    \begin{subfigure}[b]{0.55\linewidth}
    \includegraphics[width=0.49\linewidth]{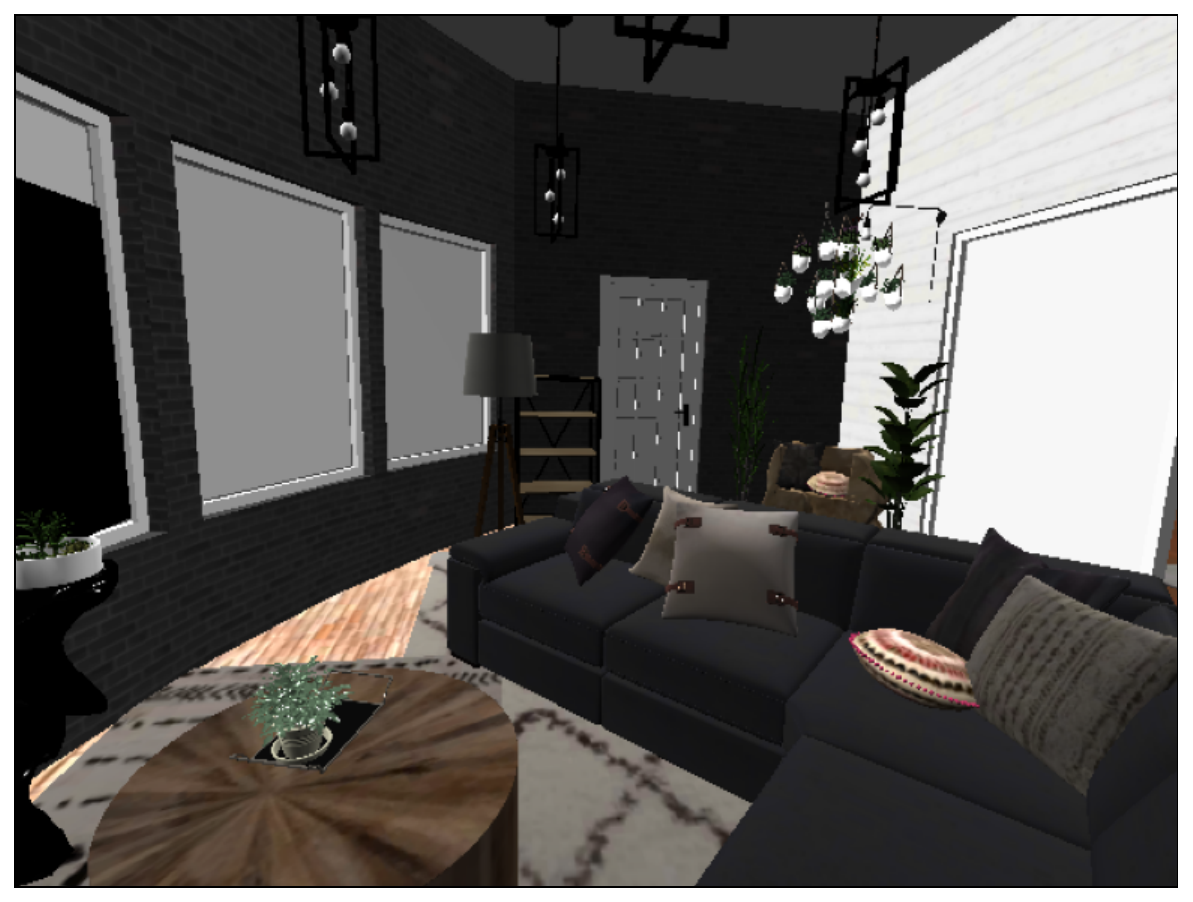}
    \includegraphics[width=0.49\linewidth]{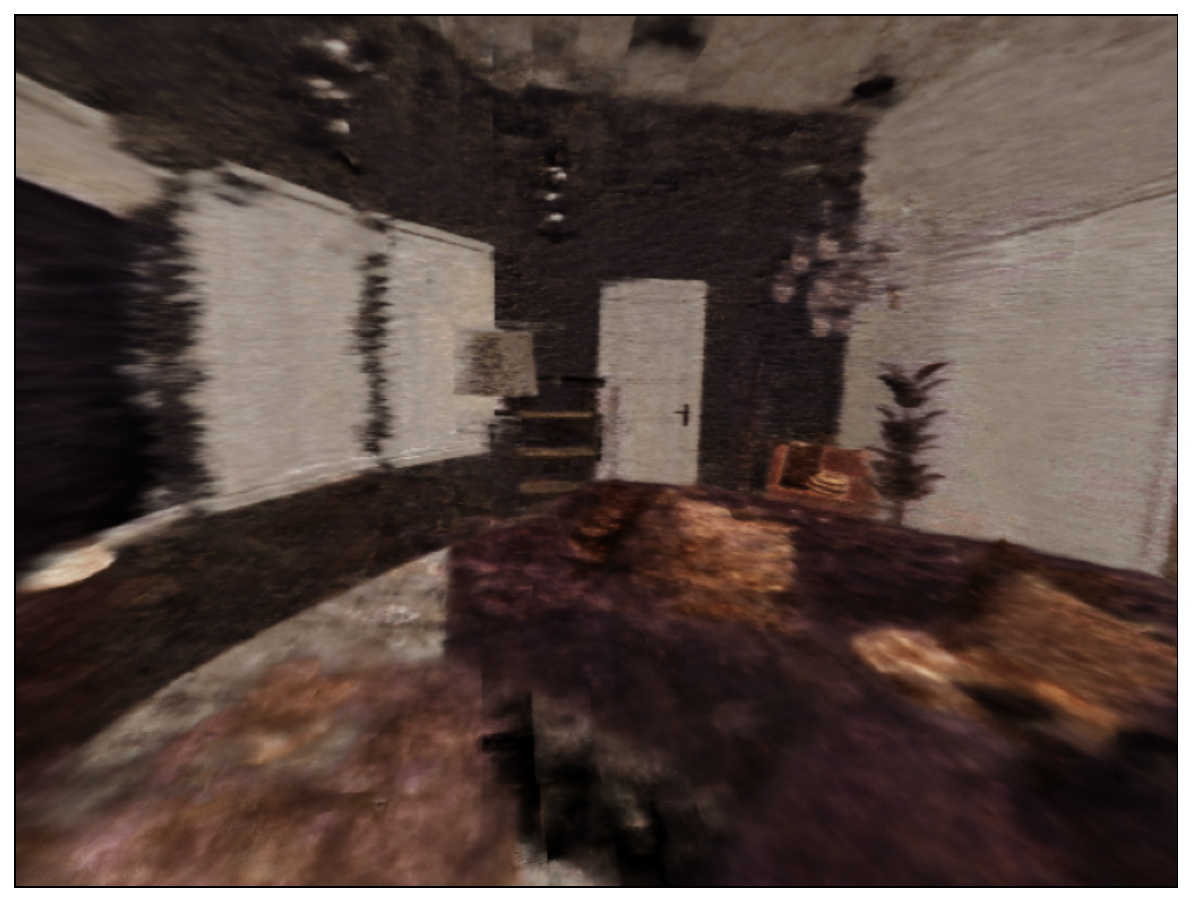}
    \includegraphics[width=0.49\linewidth]{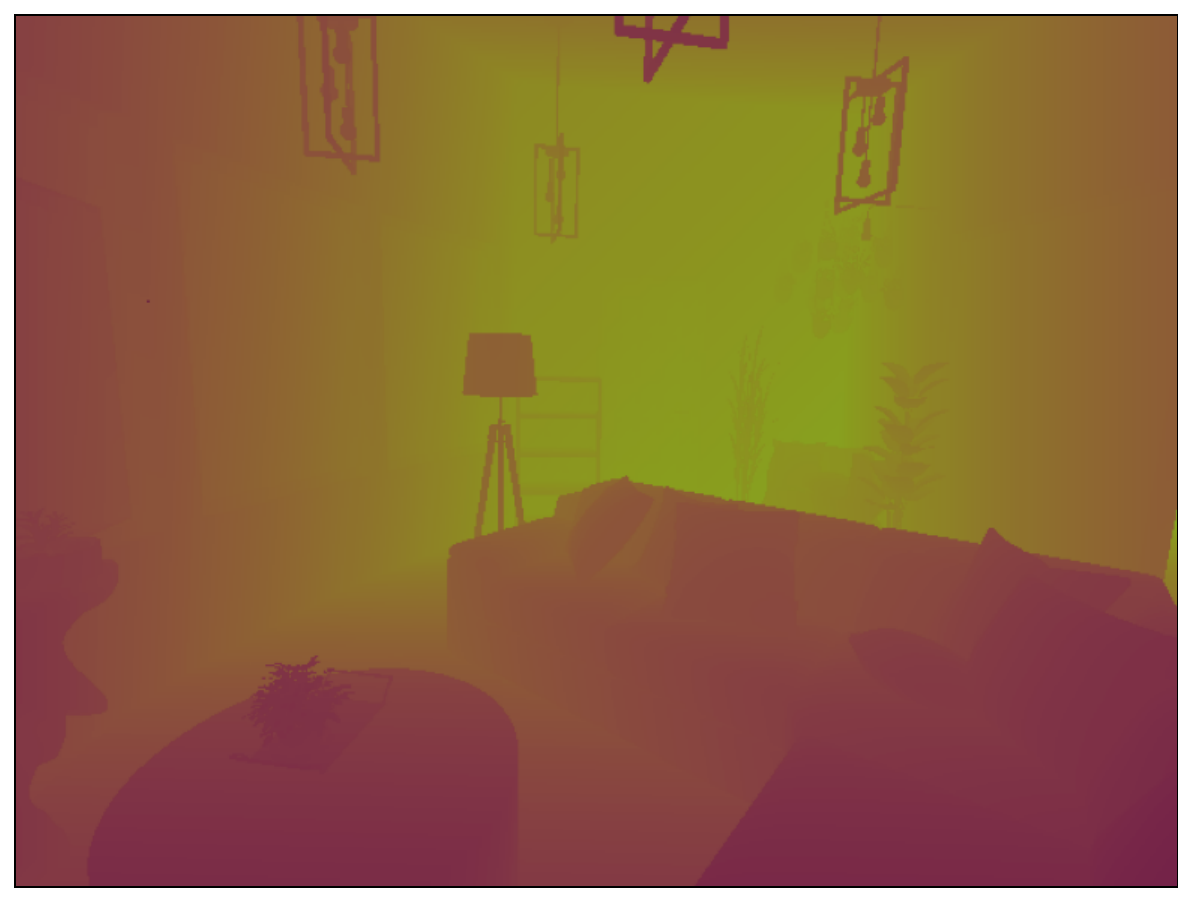}
    \includegraphics[width=0.49\linewidth]{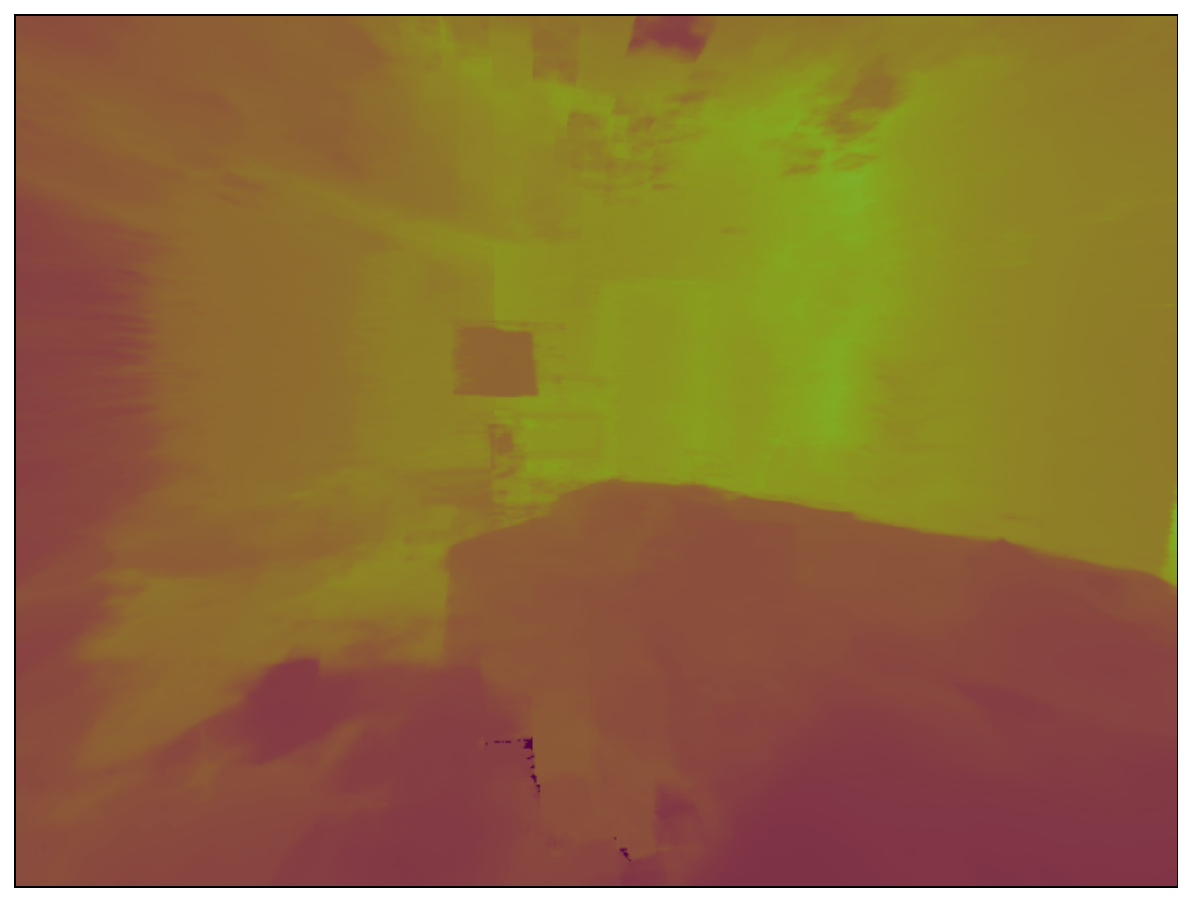}
    \includegraphics[width=0.49\linewidth]{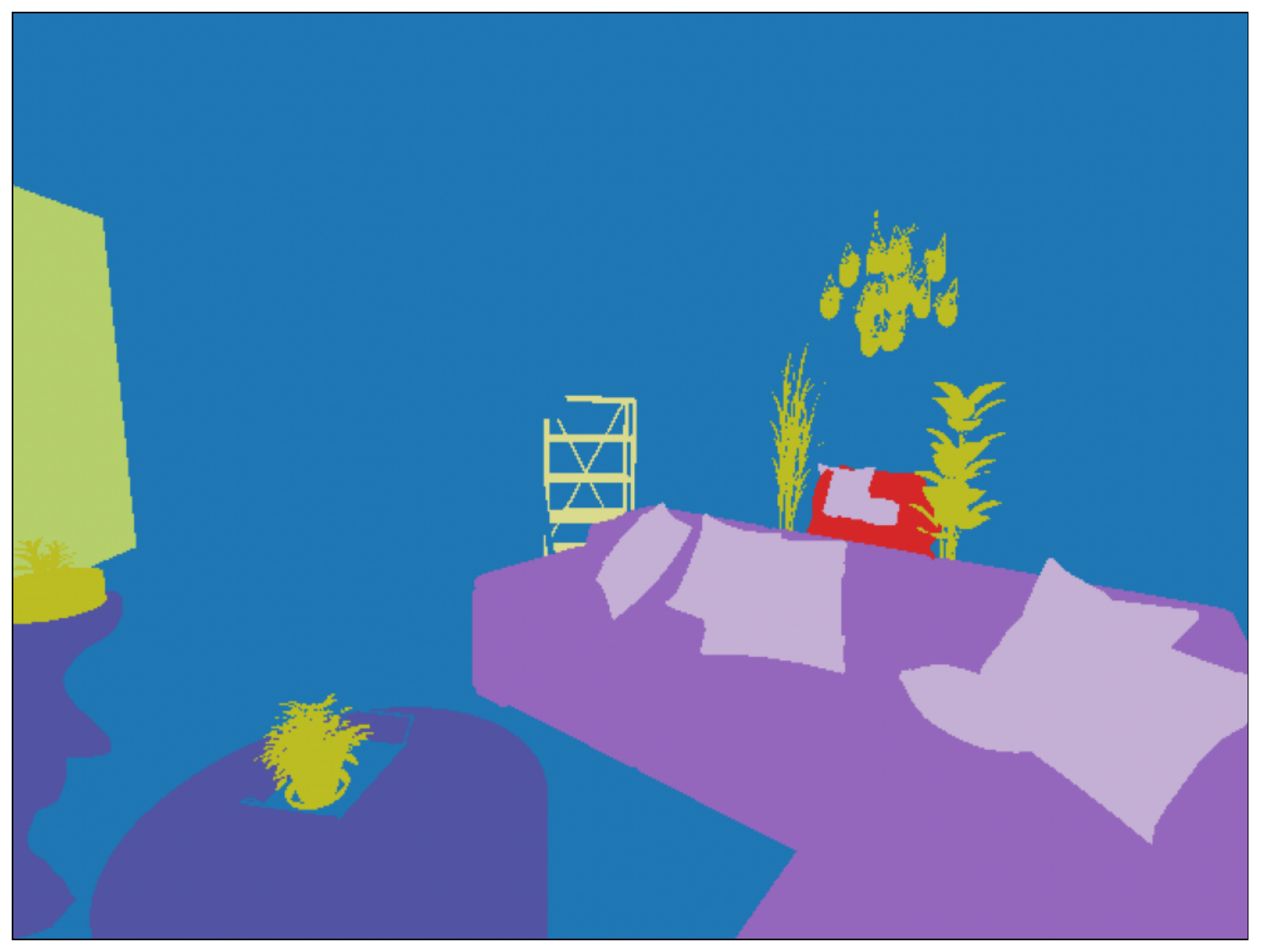}
    \includegraphics[width=0.49\linewidth]{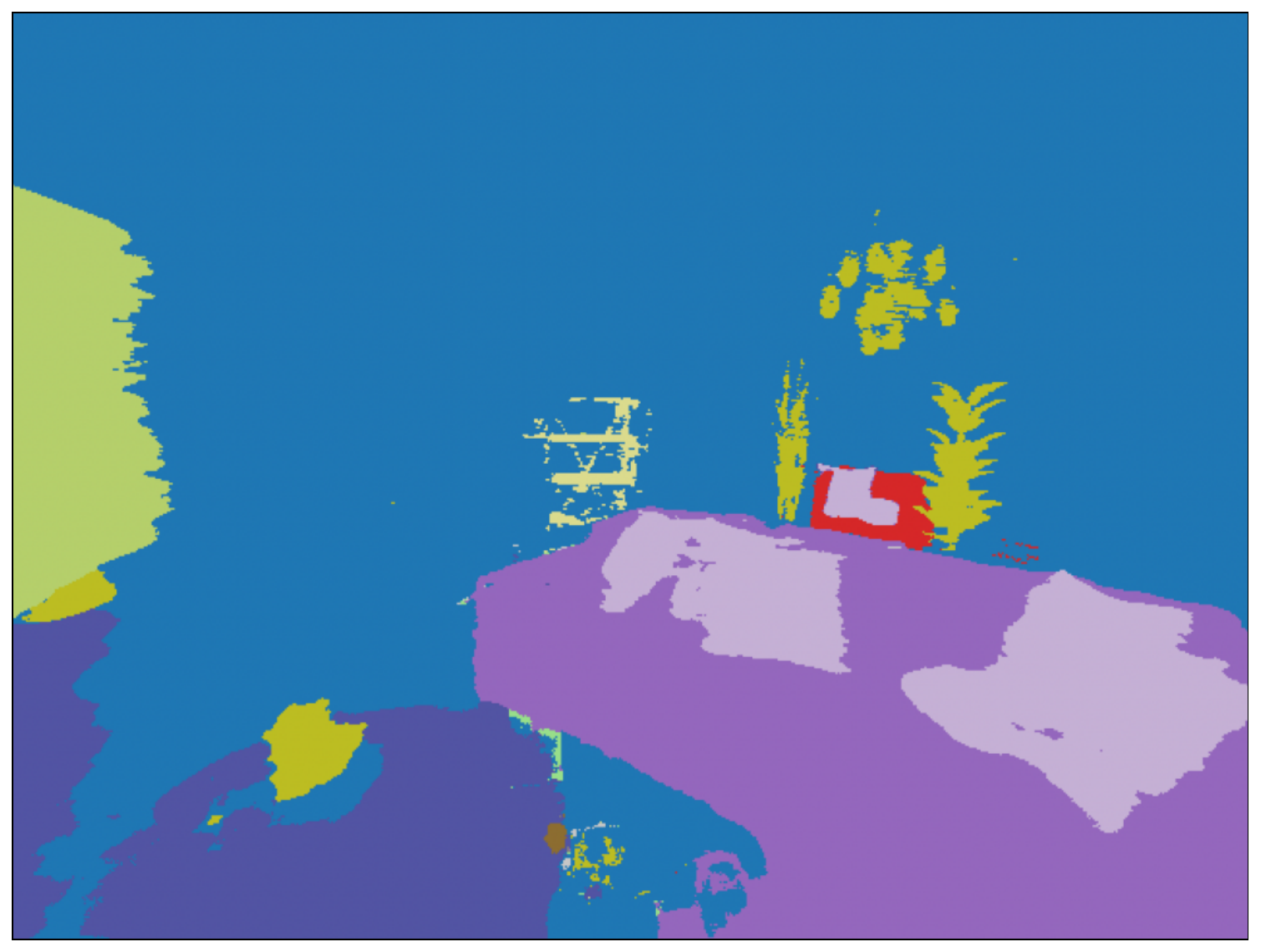}
    \end{subfigure}
    \includegraphics[width=0.42\linewidth]{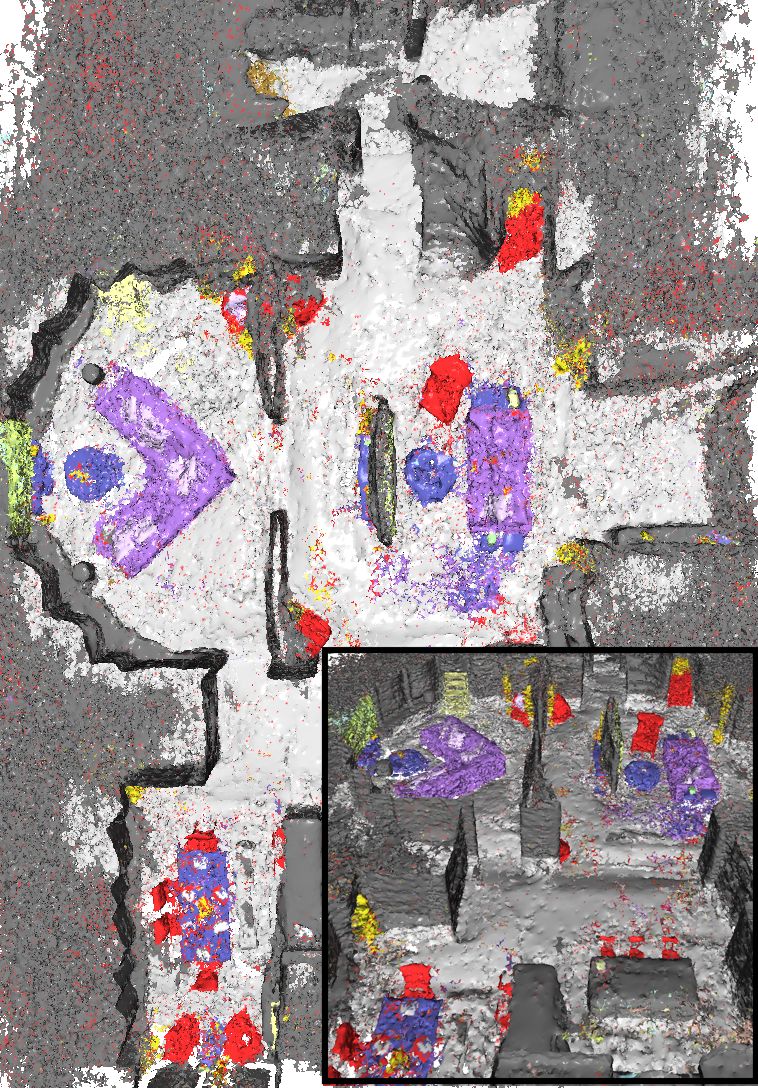}
    \caption{Ground truth observation (left) and predicted values (middle) of RGB (top), depth (middle) and semantic segmentation (bottom) after exploring scene 2; the final NeRF mesh is shown on the right.}
    \label{fig:scene2results}
\end{figure}

\cref{fig:scene3map} describes the result of exploring an unknown scene. For the sake of clarity, we draw only a 2D cross section of the 3D voxel map from the NeRF. At the start of the experiment, adjacent rooms are unexplored and hence provide high predictive information. This results in a trajectory that goes through doorways to explore new rooms. This is a sophisticated behavior and it is achieved without any need to explicitly define doorways, using just the predictive information objective. From the initial map before the quadrotor starts moving (black boundary in~\cref{fig:scene3map}), the known map expands to the one marked using the white boundaries after a single iteration. In our picture, the known free space appears disconnected because of dilation of obstacles. Synthesized views and meshes of reconstructed scene 1 and scene 2 are shown in~\cref{fig:Overview,fig:uqe,fig:scene2results}.

\begin{figure}[!ht]
    \centering
    \begin{minipage}{0.43\linewidth}
        \centering
        \includegraphics[width=\linewidth]{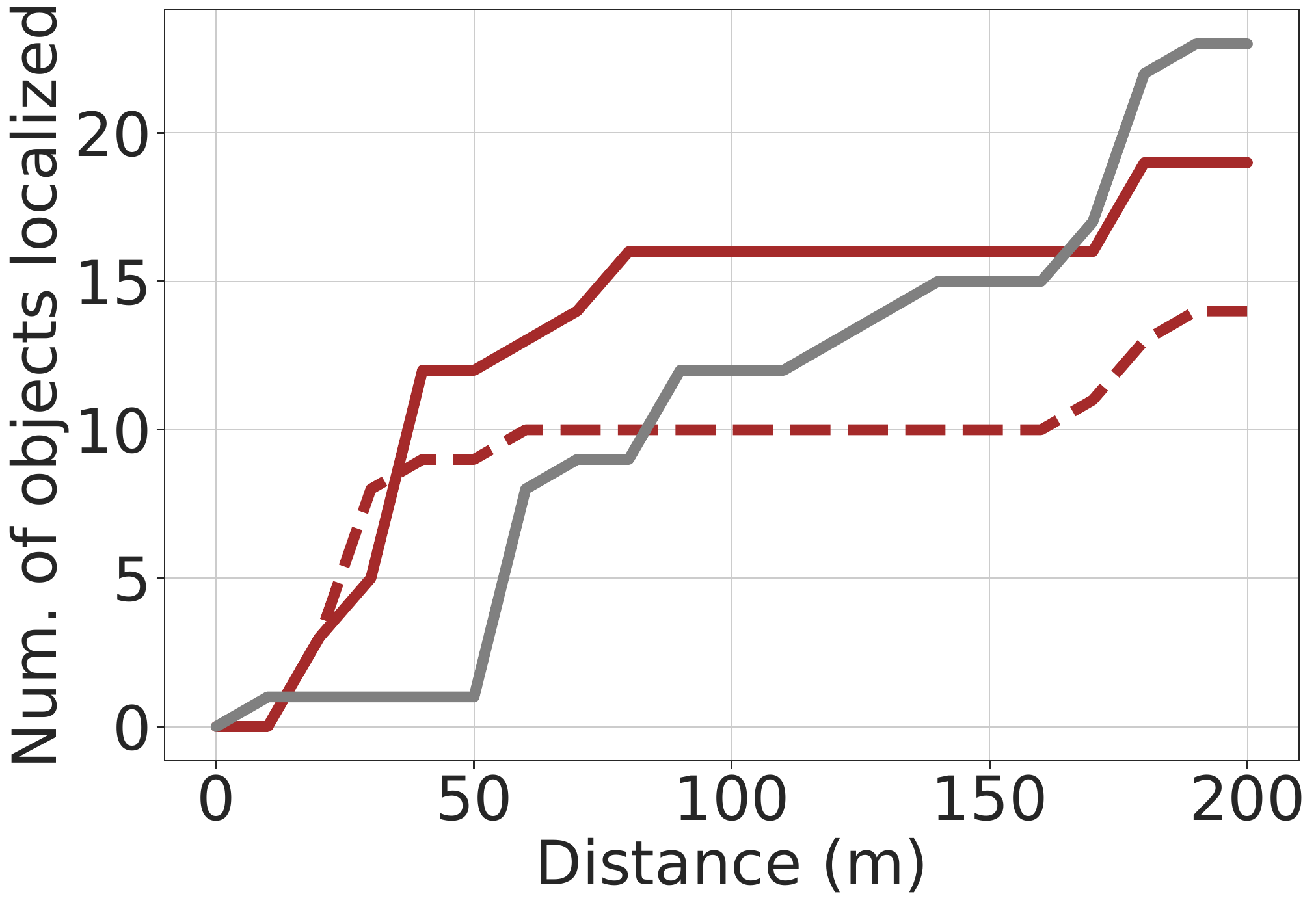}
        \includegraphics[width=\linewidth]{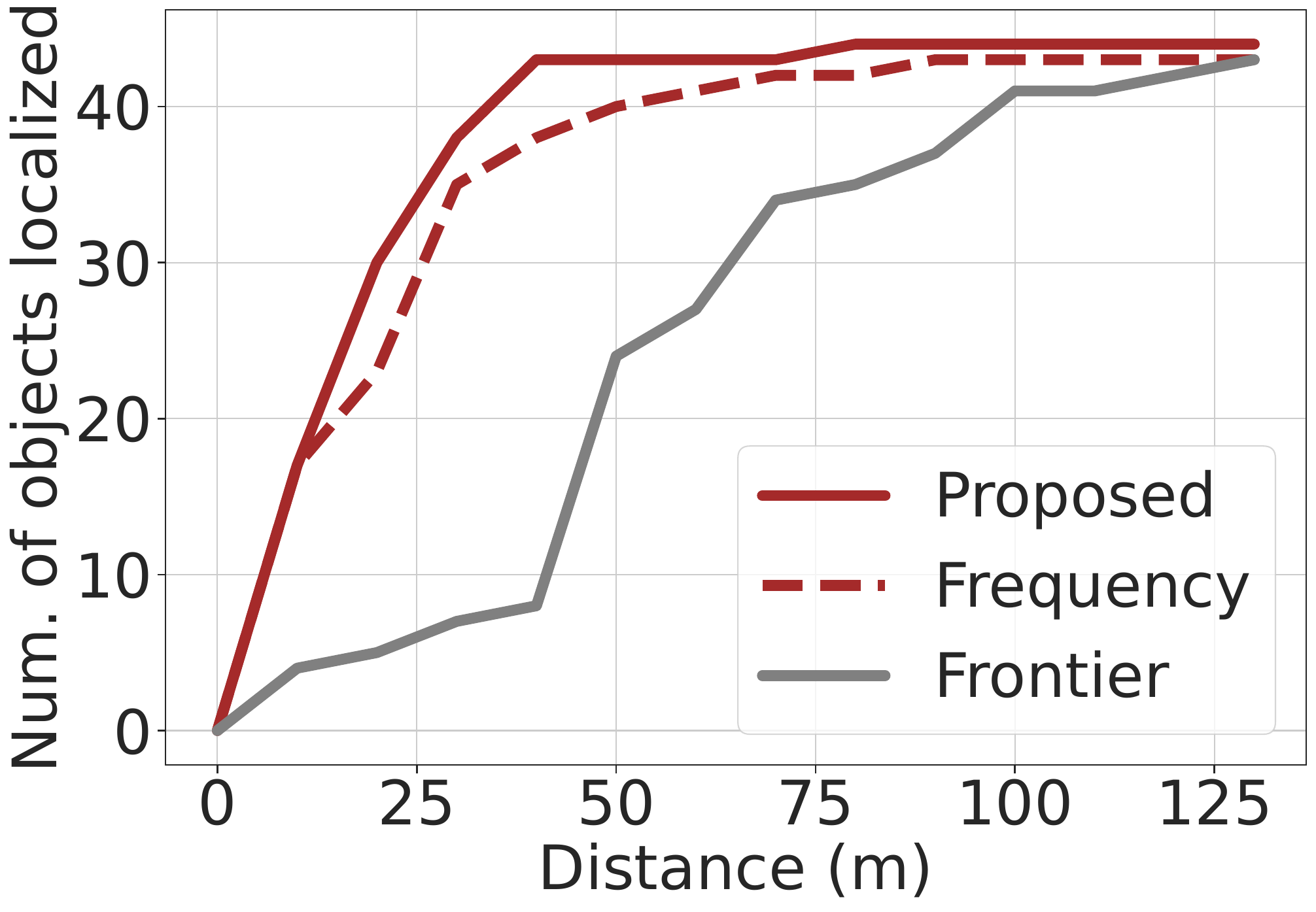}
        \includegraphics[width=\linewidth]{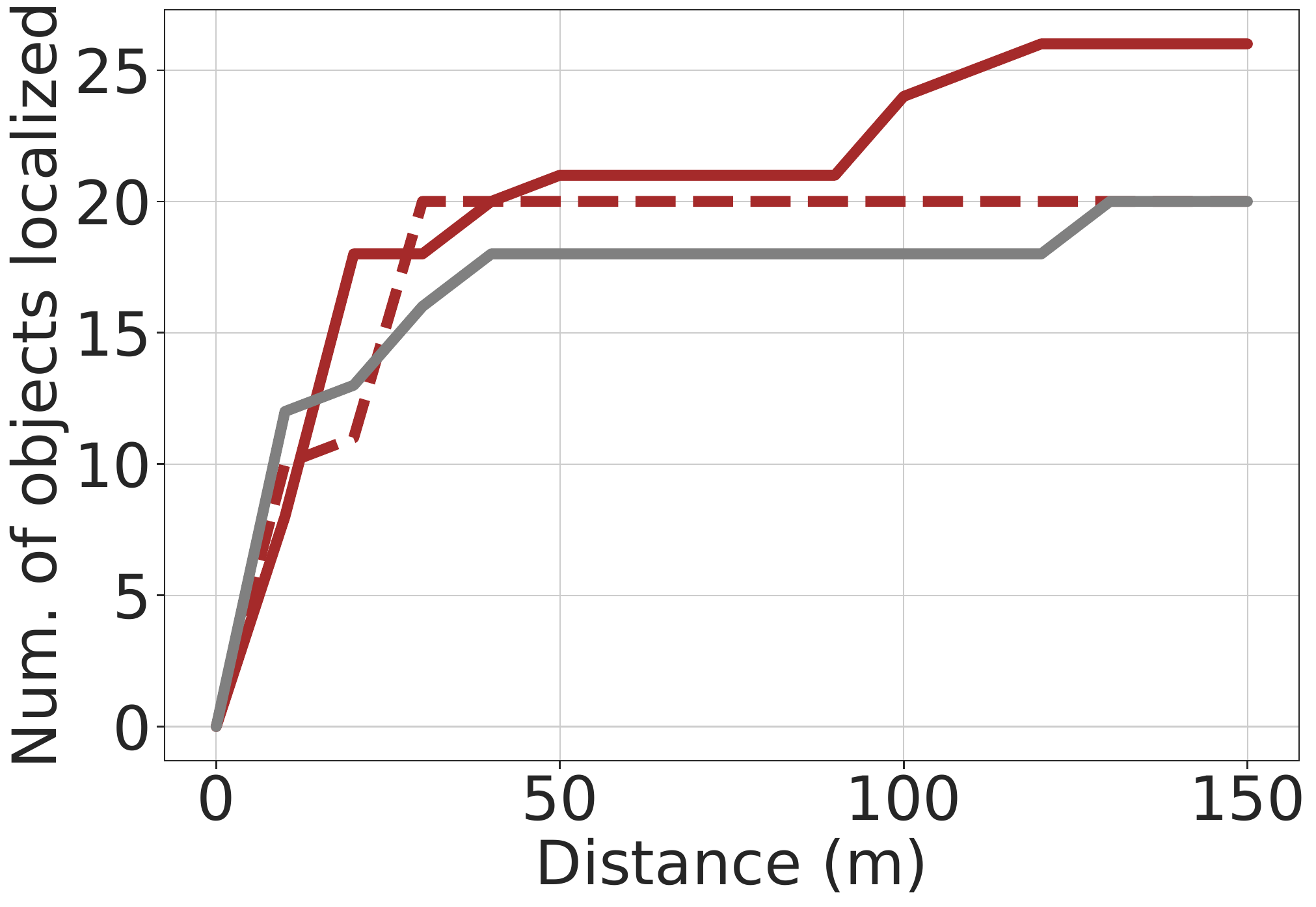}
    \end{minipage}%
    \begin{minipage}{0.55\linewidth}
        \centering
        \includegraphics[width=\linewidth]{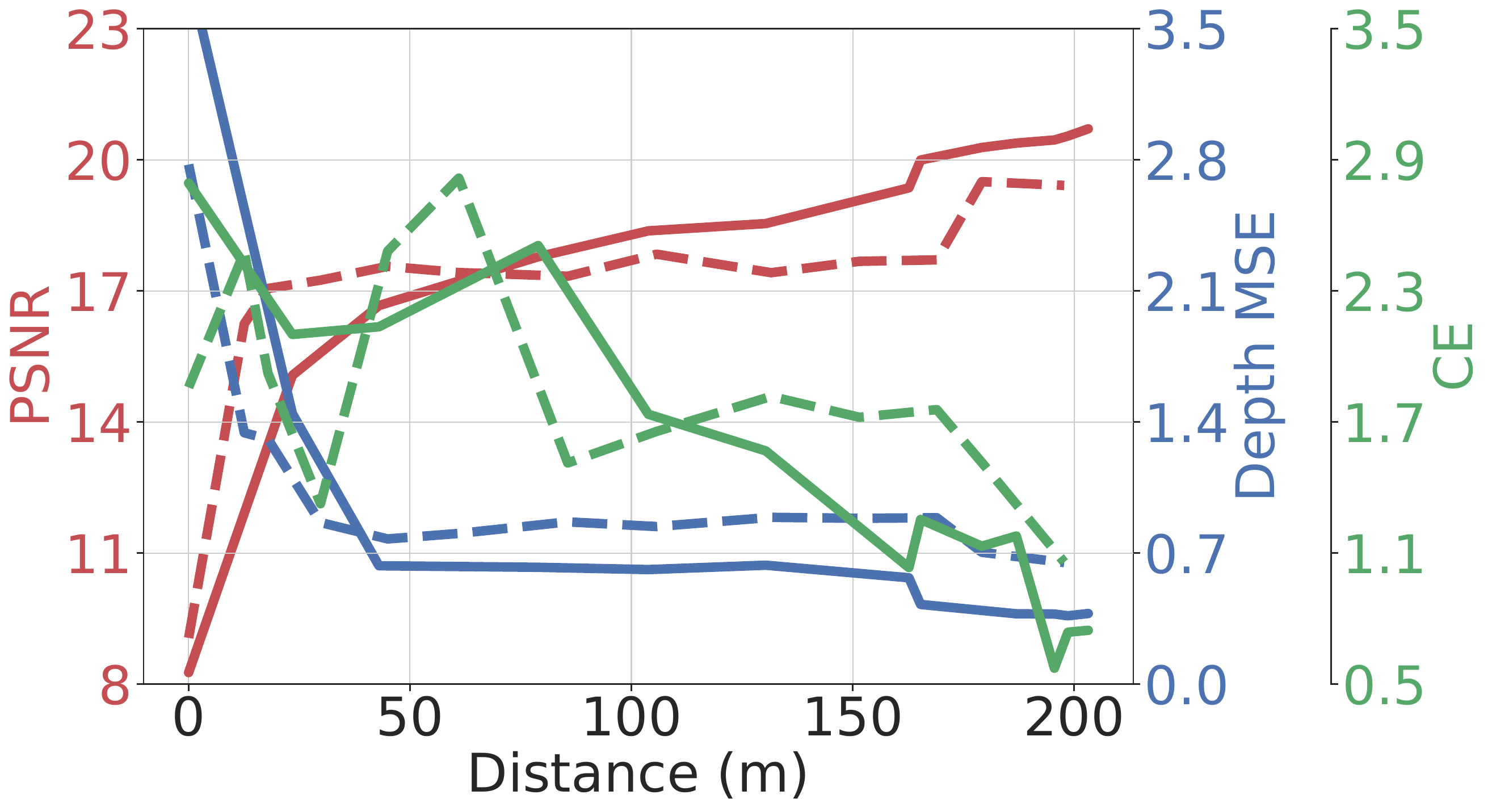}
          \includegraphics[width=\linewidth]{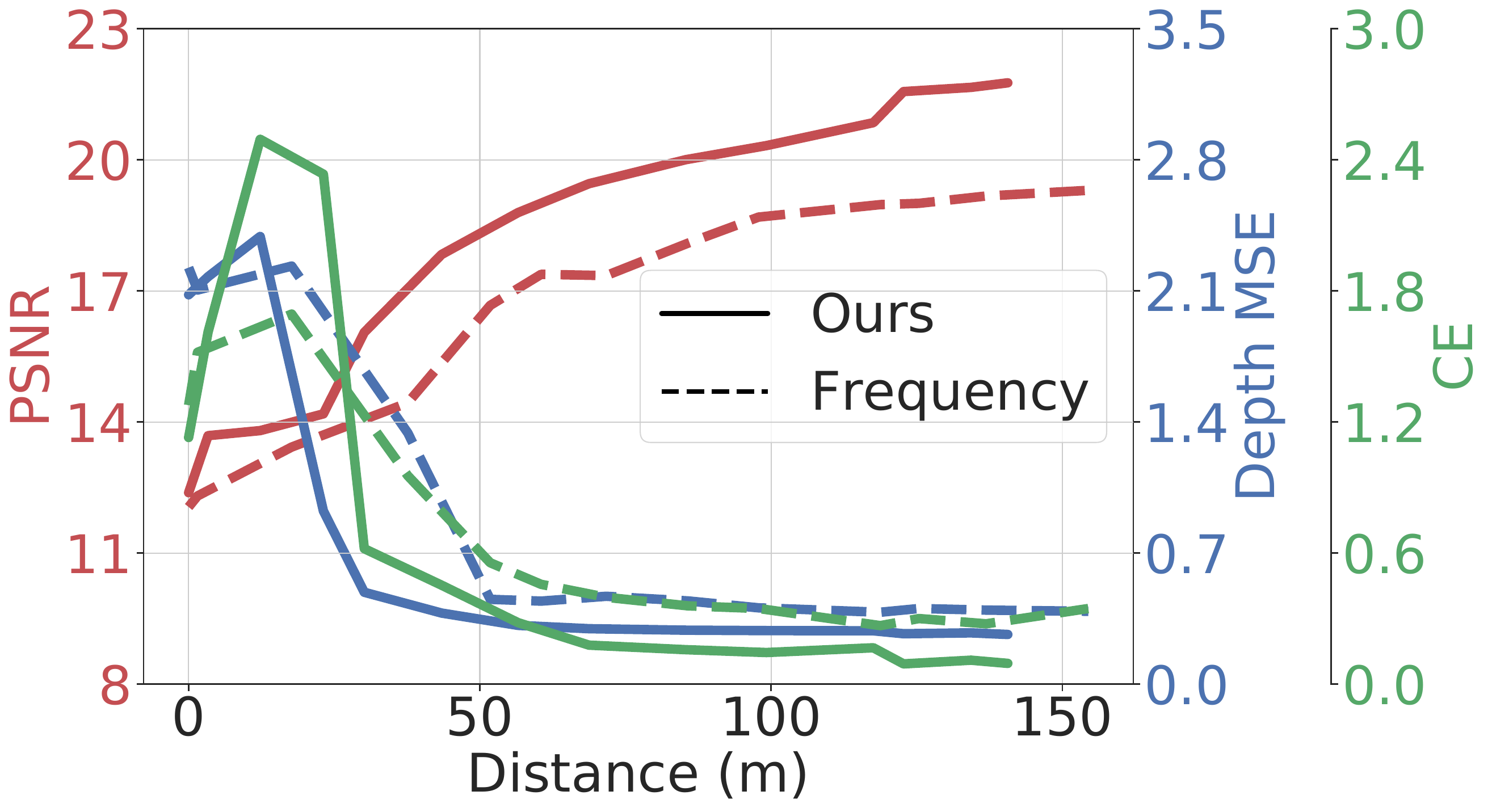}
        \includegraphics[width=\linewidth]{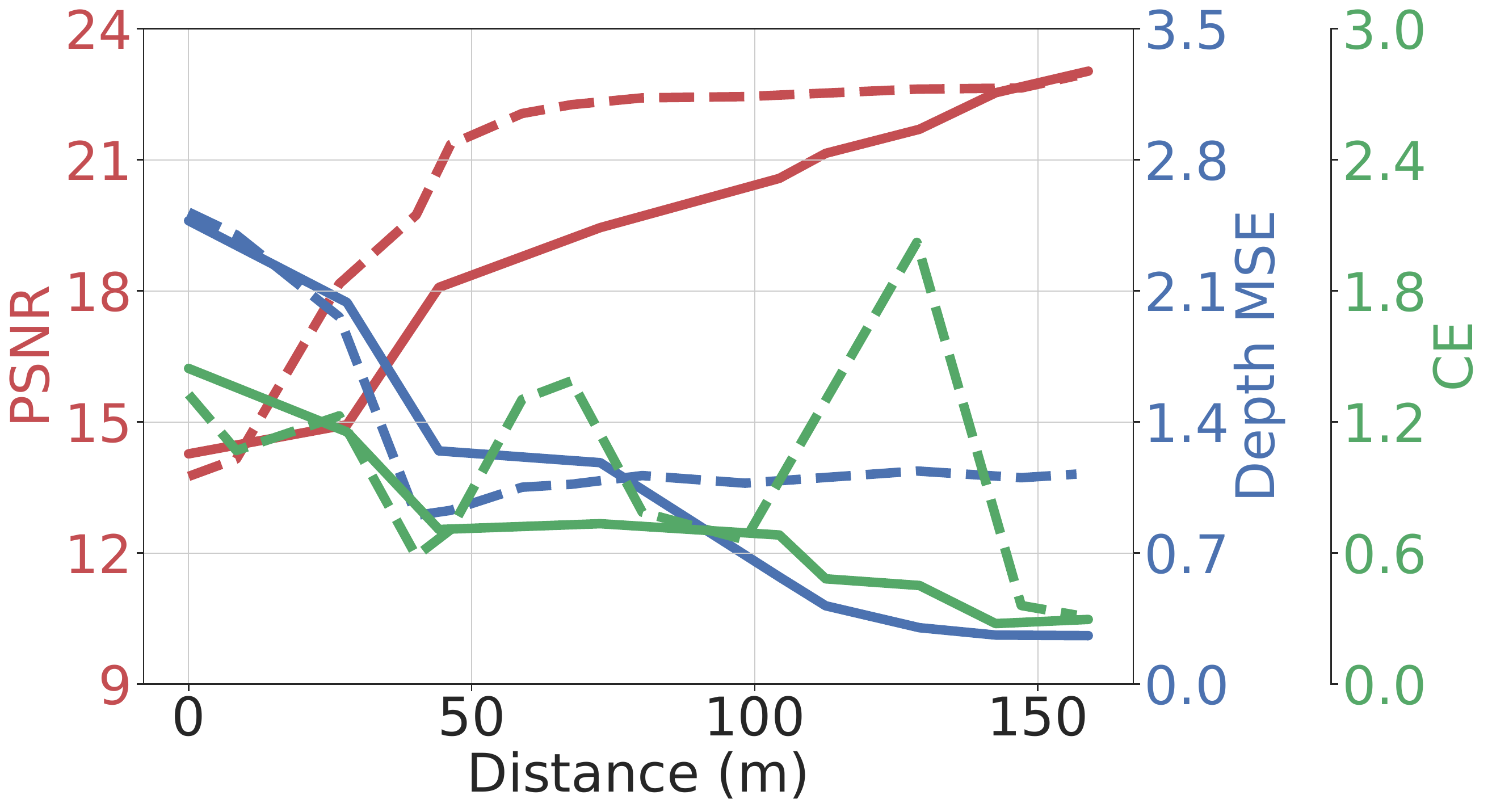}
    \end{minipage}
    \caption{\textbf{Object localization results (left) and scene reconstruction quality (right)}. Results for scene1, scene2 and scene3 are presented from top to bottom respectively. For scene reconstruction quality, four test viewpoints are chosen from each room to evaluate NeRF. There are 56, 32, 36 test viewpoints respectively. Based on our termination criteria, active perception terminates at 212m, 145m, 163m for scene 1, 2, 3 respectively. We terminate the baseline methods at approximately the same distance for comparison.}
    \label{fig:recon_qua}
\end{figure}

\cref{fig:recon_qua} (right) shows the improvement in reconstruction quality after active perception. Our method in general performs better in peak signal-to-noise ratio (PSNR) of synthesized RGB image, depth mean squared error (MSE), and cross entropy (CE) of pixel-wise classification. 

\subsection{Discussion}

\paragraph*{Exploration behavior} We observe that our method outperforms the baselines in the first half of each experiment, but that all three methods eventually obtain similar results towards the end of the experiment. The selection of trajectories to maximize predictive information in our method enables the quadrotor to quickly explore unseen or partially seen areas in the scene such as large rooms and occluded objects. This explains the sharp increase in the number of detected objects initially. However, by maximizing predictive information, our approach will eventually transition to exploiting seen areas to improve the quality of the NeRF (seeking detailed photometric and depth information), rather than prioritizing small unseen areas with relatively lower predictive information. In contrast, the frontier-based strategy will greedily move to the closest frontiers and eventually explore all regions, albeit inefficiently.

We can see from~\cref{fig:recon_qua} that our method achieves better scene reconstruction quality than the \emph{Frequency} baseline despite the random sampling of final waypoints in our method. In scene 1, the \emph{Frequency} baseline performs worse because the method will be lazy and not move into the hallway that has been observed many times. This problem can be solved by considering view-angle dependent uniform sampling~\cite{kopanas2023improving}. However, such a method also increases the sampling complexity since the agent has to view each location from a different viewpoint. This would be inefficient, e.g. viewing a plain wall from different angles wouldn't provide much more information, so it would take a longer distance for such a method to explore the whole scene.

\paragraph*{Complexity of the scene} In scene 2, our method identifies objects significantly faster than the \emph{Frontier} baseline at the beginning because scene 2 is simpler (hallway is wider, which enables us to see a lot of free space quickly). The frontier-based method prioritizes the closest frontiers and does not exploit such properties of the scene. This also occurs in the first 150 m of scene 1 when the quadrotor is exploring the large rooms with a wide doorway. By the end of the experiment in scene 1, the frontier-based method eventually achieves higher coverage by thoroughly exploring the scene through frontiers. The voxel grid obtained from the NeRF is less accurate, and the dilation of obstacles causes some narrow hallways and doors to be blocked off. In addition, since exploring small rooms might give less predictive information than revisiting larger areas, our approach may not visit such areas until larger areas are well-learned.

%% file: discussion.tex

\section{Conclusion}
\label{s:conclusion}
We argued that behaviors that resemble active perception can be developed by maximizing the predictive information, i.e., the mutual information of the past observations with future ones. Instantiating this approach requires a representation that (a) summarizes past observations, (b) synthesizes future observations, and (c) calculates predictive information along trajectories in the state-space. We conducted simulation experiments in photorealistic indoor scenes to study object localization and scene reconstruction tasks. We showed that we can faithfully instantiate this first-principles approach to active perception to build (a) high-quality maps of indoor scenes, (b) identify and localize a large number of objects. It is remarkable that a single principle, without any additional heuristics, can produce sophisticated behaviors that are often hand-engineered and tuned to specific problems. We believe that the practical performance of this approach can be further improved by combining existing approaches to active perception with our information-theoretic formulation.

%% file: acknowledgments.tex

\section{Acknowledgments}
\label{s:acknowledgments}
We gratefully acknowledge the support of NIFA grant 2022-67021-36856, NSF grants 2112665, IIS-2145164, and the IoT4Ag ERC funded by the National Science Foundation grant EEC-1941529. Siming He acknowledges the support of the Wharton Summer Program for Undergraduate Research. Dexter Ong acknowledges the support of DSO National Laboratories, Singapore.

%% file: main.bbl
\begin{thebibliography}{10}

\bibitem{bajcsy2016revisiting}
R.~Bajcsy, Y.~Aloimonos, and J.~K. Tsotsos, ``Revisiting active perception,'' 2016.

\bibitem{Aloimonos2004ActiveV}
Y.~Aloimonos, I.~Weiss, and A.~Bandyopadhyay, ``Active vision,'' {\em International Journal of Computer Vision}, vol.~1, pp.~333--356, 2004.

\bibitem{Tsotsos2007AttentionAV}
J.~K. Tsotsos and K.~Shubina, ``Attention and visual search: Active robotic vision systems that search,'' 2007.

\bibitem{scaramuzza2017quadactivevision}
D.~Falanga, E.~Mueggler, M.~Faessler, and D.~Scaramuzza, ``Aggressive quadrotor flight through narrow gaps with onboard sensing and computing using active vision,'' in {\em ICRA}, pp.~5774--5781, 2017.

\bibitem{novkovic2020object}
T.~Novkovic, R.~Pautrat, F.~Furrer, M.~Breyer, R.~Siegwart, and J.~Nieto, ``Object finding in cluttered scenes using interactive perception,'' 2020.

\bibitem{dosovitskiy2016learning}
A.~Dosovitskiy and V.~Koltun, ``Learning to act by predicting the future,'' {\em arXiv preprint arXiv:1611.01779}, 2016.

\bibitem{mildenhall2020nerf}
B.~Mildenhall, P.~P. Srinivasan, M.~Tancik, J.~T. Barron, R.~Ramamoorthi, and R.~Ng, ``Nerf: {{Representing}} scenes as neural radiance fields for view synthesis,'' in {\em European Conference on Computer Vision}, pp.~405--421, 2020.

\bibitem{inplaceICCV2021}
S.~Zhi, T.~Laidlow, S.~Leutenegger, and A.~Davison, ``In-place scene labelling and understanding with implicit scene representation,'' in {\em ICCV}, 2021.

\bibitem{mueller2022instant}
T.~M\"uller, A.~Evans, C.~Schied, and A.~Keller, ``Instant neural graphics primitives with a multiresolution hash encoding,'' {\em ACM Trans. Graph.}, vol.~41, pp.~102:1--102:15, July 2022.

\bibitem{li2023nerfacc}
R.~Li, H.~Gao, M.~Tancik, and A.~Kanazawa, ``{NerfAcc}: Efficient sampling accelerates {NeRFs},'' 2023.

\bibitem{bialek2001predictability}
W.~Bialek, I.~Nemenman, and N.~Tishby, ``Predictability, complexity, and learning,'' {\em Neural computation}, 2001.

\bibitem{pan2022activenerf}
X.~Pan, Z.~Lai, S.~Song, and G.~Huang, ``Activenerf: Learning where to see with uncertainty estimation,'' 2022.

\bibitem{smith2022uncertaintydriven}
E.~J. Smith, M.~Drozdzal, D.~Nowrouzezahrai, D.~Meger, and A.~Romero-Soriano, ``Uncertainty-driven active vision for implicit scene reconstruction,'' 2022.

\bibitem{yamauchi1997frontier}
B.~Yamauchi, ``A frontier-based approach for autonomous exploration,'' in {\em CIRA}, pp.~146--151, 1997.

\bibitem{chaplot2020learning}
D.~S. Chaplot, D.~Gandhi, S.~Gupta, A.~Gupta, and R.~Salakhutdinov, ``Learning to explore using active neural slam,'' 2020.

\bibitem{kopanas2023improving}
G.~Kopanas and G.~Drettakis, ``Improving {NeRF} quality by progressive camera placement for navigation in complex environments,'' 2023.

\bibitem{marza2023autonerf}
P.~Marza, L.~Matignon, O.~Simonin, D.~Batra, C.~Wolf, and D.~S. Chaplot, ``Autonerf: Training implicit scene representations with autonomous agents,'' 2023.

\bibitem{adamkiewicz2022visiononly}
M.~Adamkiewicz, T.~Chen, A.~Caccavale, R.~Gardner, P.~Culbertson, J.~Bohg, and M.~Schwager, ``Vision-only robot navigation in a neural radiance world,'' 2022.

\bibitem{zhan2022activermap}
H.~Zhan, J.~Zheng, Y.~Xu, I.~Reid, and H.~Rezatofighi, ``Activermap: Radiance field for active mapping and planning,'' 2022.

\bibitem{Ran_2023}
Y.~Ran, J.~Zeng, S.~He, J.~Chen, L.~Li, Y.~Chen, G.~Lee, and Q.~Ye, ``{NeurAR}: Neural uncertainty for autonomous 3d reconstruction with implicit neural representations,'' {\em {IEEE} Robotics and Automation Letters}, vol.~8, pp.~1125--1132, feb 2023.

\bibitem{shen2021stochastic}
J.~Shen, A.~Ruiz, A.~Agudo, and F.~Moreno-Noguer, ``Stochastic neural radiance fields: Quantifying uncertainty in implicit {3D} representations,'' 2021.

\bibitem{jiang2023fisherrf}
W.~Jiang, B.~Lei, and K.~Daniilidis, ``Fisherrf: Active view selection and uncertainty quantification for radiance fields using fisher information,'' 2023.

\bibitem{sunderhauf2022densityaware}
N.~Sunderhauf, J.~Abou-Chakra, and D.~Miller, ``Density-aware {NeRF} ensembles: Quantifying predictive uncertainty in neural radiance fields,'' 2022.

\bibitem{cover1999elements}
T.~M. Cover, {\em Elements of Information Theory}.
\newblock 1999.

\bibitem{vaswani2017attention}
A.~Vaswani, N.~Shazeer, N.~Parmar, J.~Uszkoreit, L.~Jones, A.~N. Gomez, L.~Kaiser, and I.~Polosukhin, ``Attention is all you need,'' in {\em Advances in Neural Information Processing Systems}, 2017.

\bibitem{yu2023NeRFbridge}
J.~Yu, J.~E. Low, K.~Nagami, and M.~Schwager, ``Nerfbridge: Bringing real-time, online neural radiance field training to robotics,'' 2023.

\bibitem{MellingerMinSnap}
D.~Mellinger and V.~Kumar, ``Minimum snap trajectory generation and control for quadrotors,'' in {\em ICRA}, pp.~2520--2525, 2011.

\bibitem{folk2023rotorpy}
S.~Folk, J.~Paulos, and V.~Kumar, ``Rotorpy: A python-based multirotor simulator with aerodynamics for education and research,'' {\em arXiv preprint arXiv:2306.04485}, 2023.

\bibitem{sakai2018pythonrobotics}
A.~Sakai, D.~Ingram, J.~Dinius, K.~Chawla, A.~Raffin, and A.~Paques, ``Pythonrobotics: a python code collection of robotics algorithms,'' 2018.

\bibitem{szot2021habitat}
A.~Szot, A.~Clegg, E.~Undersander, E.~Wijmans, Y.~Zhao, J.~Turner, N.~Maestre, M.~Mukadam, D.~Chaplot, O.~Maksymets, A.~Gokaslan, V.~Vondrus, S.~Dharur, F.~Meier, W.~Galuba, A.~Chang, Z.~Kira, V.~Koltun, J.~Malik, M.~Savva, and D.~Batra, ``Habitat 2.0: Training home assistants to rearrange their habitat,'' in {\em Advances in Neural Information Processing Systems (NeurIPS)}, 2021.

\bibitem{khanna2023hssd}
M.~{Khanna*}, Y.~{Mao*}, H.~Jiang, S.~Haresh, B.~Shacklett, D.~Batra, A.~Clegg, E.~Undersander, A.~X. Chang, and M.~Savva, ``{Habitat Synthetic Scenes Dataset: An Analysis of 3D Scene Scale and Realism Tradeoffs for ObjectGoal Navigation},'' {\em arXiv preprint}, 2023.

\end{thebibliography}
